\documentclass[sigconf,usenames,dvipsnames]{acmart}

\AtBeginDocument{
  \providecommand\BibTeX{{
    \normalfont B\kern-0.5em{\scshape i\kern-0.25em b}\kern-0.8em\TeX}}}

\copyrightyear{2020}
\acmYear{2020}
\setcopyright{rightsretained}
\acmConference[KDD '20]{Proceedings of the 26th ACM SIGKDD Conference on Knowledge Discovery and Data Mining}{August 23--27, 2020}{Virtual Event, CA, USA}
\acmBooktitle{Proceedings of the 26th ACM SIGKDD Conference on Knowledge Discovery and Data Mining (KDD '20), August 23--27, 2020, Virtual Event, CA, USA}
\acmDOI{10.1145/3394486.3403093}
\acmISBN{978-1-4503-7998-4/20/08}

\usepackage{booktabs} 
\usepackage{color,colortbl}
\usepackage{xcolor}
\usepackage{amsmath}
\usepackage{amsfonts}
\usepackage{amssymb}
\usepackage{mathtools}
\usepackage{bm}
\usepackage{graphicx}
\usepackage{paralist}
\usepackage{enumitem}
\setlist[itemize]{nosep,leftmargin=1em}  
\setlist[enumerate]{nosep,leftmargin=1.5em}  

\usepackage{xspace}
\usepackage[labelformat=simple]{subcaption}

\usepackage{makecell}
\usepackage[vlined,ruled,linesnumbered]{algorithm2e}
\usepackage{hyperref}
\usepackage{cleveref}
\usepackage{multirow}
\usepackage[breakwords]{truncate}
\usepackage{pdfrender}
\usepackage[most]{tcolorbox}

\definecolor{Gray}{gray}{0.9}
\definecolor{LightGray}{gray}{0.96}
\definecolor{LightCyan}{rgb}{0.88,1,1}
\newcommand{\best}[1]{\textbf{\cellcolor{Gray}#1}}
\newcommand{\secondbest}[1]{\underline{\cellcolor{LightGray}#1}}

\newcommand{\tablefont}{}

\newcommand*{\belowrulesepcolor}[1]{
	\noalign{
		\kern-\belowrulesep
		\begingroup
		\color{#1}
		\hrule height\belowrulesep
		\endgroup
	}
}
\newcommand*{\aboverulesepcolor}[1]{
	\noalign{
		\begingroup
		\color{#1}
		\hrule height\aboverulesep
		\endgroup
		\kern-\aboverulesep
	}
}

\newcommand*{\boldcheckmark}{
	\textpdfrender{
		TextRenderingMode=FillStroke,
		LineWidth=.7pt, 
	}{\checkmark}
}

\newcommand{\sgnl}{signal\xspace}

\newcommand{\Sgnls}{Signals\xspace}
\newcommand{\sgnls}{signals\xspace}
\newcommand{\signal}{input signal\xspace}
\newcommand{\Signal}{Input signal\xspace}
\newcommand{\SIGNAL}{Input Signal\xspace}
\newcommand{\signals}{input signals\xspace}
\newcommand{\Signals}{Input signals\xspace}
\newcommand{\SIGNALS}{Input Signals\xspace}

\newcommand{\rebel}{rebel\xspace}
\newcommand{\Rebel}{Rebel\xspace}

\newcommand{\vect}[1]{\bm{\mathbf{#1}}}
\newcommand{\mat}[1]{\bm{\mathbf{\MakeUppercase{#1}}}}

\newcommand{\kgtensor}{{\bm{G}}} 
\newcommand{\nodefeat}{{\bm{X}}} 
\newcommand{\kg}{{\bm{\mathcal{G}}}} 
\newcommand{\modelparams}{{\bm{\theta}}} 
\newcommand{\setS}{{\bm{S}}} 
\newcommand{\s}{{\bm{s}}} 
\newcommand{\numS}{{M}} 
 
\newcommand{\z}{{\bm{z}}} 
\newcommand{\N}{{\mathcal{N}}}

\newcommand{\tmdb}{\textsc{tmdb5k}\xspace}
\newcommand{\music}{\textsc{music10k}\xspace}
\newcommand{\fb}{\textsc{fb15k}\xspace}
\newcommand{\imdb}{\textsc{imdb}\xspace}

\newcommand{\method}{\textsc{Multi\-Import}\xspace}
\newcommand{\methodone}{\textsc{Multi\-Import-1}\xspace}

\newcommand{\geni}{\textsc{GENI}\xspace}
\newcommand{\pr}{\textsc{PR}\xspace}
\newcommand{\ppr}{\textsc{PPR}\xspace}
\newcommand{\har}{\textsc{HAR}\xspace}

\newcommand{\ndcg}{\textsc{NDCG@100}\xspace}

\newcommand{\diag}{\mathop{\mathrm{diag}}}

\begin{document}

\fancyhead{}
\setlength{\abovedisplayskip}{2pt}
\setlength{\belowdisplayskip}{2pt}
\setlength{\textfloatsep}{0.1cm}
\setlength{\floatsep}{0.1cm}

\title[\method: Inferring Node Importance from Multiple \Signals in a Knowledge Graph]{\method: Inferring Node Importance in\\ a Knowledge Graph from Multiple \SIGNALS}

\author{Namyong Park$^{1*}$, Andrey Kan$^2$, Xin Luna Dong$^2$, Tong Zhao$^2$, Christos Faloutsos$^{1*}$}
\thanks{${}^*$Work performed while at Amazon.}
\email{{namyongp,christos}@cs.cmu.edu,{avkan,lunadong,zhaoton}@amazon.com}
\affiliation{\institution{\textsuperscript{1}Carnegie Mellon University, \textsuperscript{2}Amazon}}

\renewcommand{\shortauthors}{N. Park et al.}

\begin{abstract}
Given multiple \signals, how can we infer node importance in a knowledge graph (KG)?
Node importance estimation is a crucial and challenging task 
	that can benefit a lot of applications including recommendation, search, and query disambiguation. 
A key challenge towards this goal is how to effectively use input from different sources.
On the one hand, a KG is a rich source of information, with multiple types of nodes and edges. 
On the other hand, there are external \signals, such as the number of votes or pageviews, 
which can directly tell us about the importance of entities in a KG. 
While several methods have been developed to tackle this problem,
their use of these external \sgnls has been limited as they are not designed to consider multiple \sgnls simultaneously.
In this paper, we develop an end-to-end model \method, which infers latent node importance 
from multiple, potentially overlapping, \signals.
\method is a latent variable model that captures the relation between node importance and \signals, and 
effectively learns from multiple \sgnls with potential conflicts.
Also, \method provides an effective estimator based on attentive graph neural networks.
We ran experiments on real-world KGs to show that \method handles several challenges 
involved with inferring node importance from multiple \signals, 
and consistently outperforms existing methods,
achieving up to 23.7\% higher NDCG@100 than the state-of-the-art method.
\end{abstract}



\maketitle

\vspace{-0.5em}

{\fontsize{8pt}{8pt} \selectfont
\textbf{ACM Reference Format:}\\
Namyong Park, Andrey Kan, Xin Luna Dong, Tong Zhao, Christos Faloutsos. 2020. \method: Inferring Node Importance from Multiple \Signals in a Knowledge Graph. In \textit{Proceedings of the 26th ACM SIGKDD Conference on Knowledge Discovery and Data Mining (KDD '20), August 23--27, 2020, Virtual Event, CA, USA.} ACM, New York, NY, USA, 10 pages. https://doi.org/10.1145/3394486.3403093}

\vspace{-0.5em}

\section{Introduction}
\label{sec:intro}
\begin{table}[!t]
	\setlength{\tabcolsep}{1.6mm}
	\renewcommand{\aboverulesep}{0pt}
	\renewcommand{\belowrulesep}{0pt}
	\setlength{\abovecaptionskip}{1pt}
	\tablefont
	\centering
	\caption{Comparison of \method and baselines in terms of what type of input each method considers to measure node importance in a knowledge graph. Only \method can consider multiple \signals.
	}
	\centering
		\begin{tabular}{ c | c | c | c | c | c }
			\toprule
			Input & \makecell{\textbf{\textsc{Multi}}\\\textbf{\textsc{Import}}} & \makecell{\textbf{\geni}\\\cite{DBLP:conf/kdd/ParkKDZF19}} & \makecell{\textbf{\har}\\\cite{DBLP:conf/sdm/LiNY12}} & \makecell{\textbf{\ppr}\\\cite{DBLP:conf/www/Haveliwala02}} & \makecell{\textbf{\pr}\\\cite{page1999pagerank}} \\
			\midrule
			\belowrulesepcolor{Gray}
			\textit{Graph Structure} 		& $\boldcheckmark$ & $\checkmark$ & $\checkmark$ & $\checkmark$ & $\checkmark$ \\
			\rowcolor{Gray}
			\textit{Multiple Predicates} 				& $\boldcheckmark$ & $\checkmark$ & $\checkmark$ & & \\
			\textit{Single \SIGNAL} 	& $\boldcheckmark$ & $\checkmark$ & $\checkmark$ & $ \checkmark $ & \\ \rowcolor{Gray}
			\textit{Multiple \SIGNALS}	& $\boldcheckmark$ & & & & \\
			\aboverulesepcolor{Gray}
			\bottomrule
		\end{tabular}
	\label{tab:method_comparison}
\end{table}

\newsavebox{\bigimage}
\begin{figure*}[!t]
	\par\vspace{-1.1em}\par
	\centering
	\sbox{\bigimage}{%
		\begin{subfigure}[b]{.33\textwidth}
			\centering
			\includegraphics[width=\textwidth]{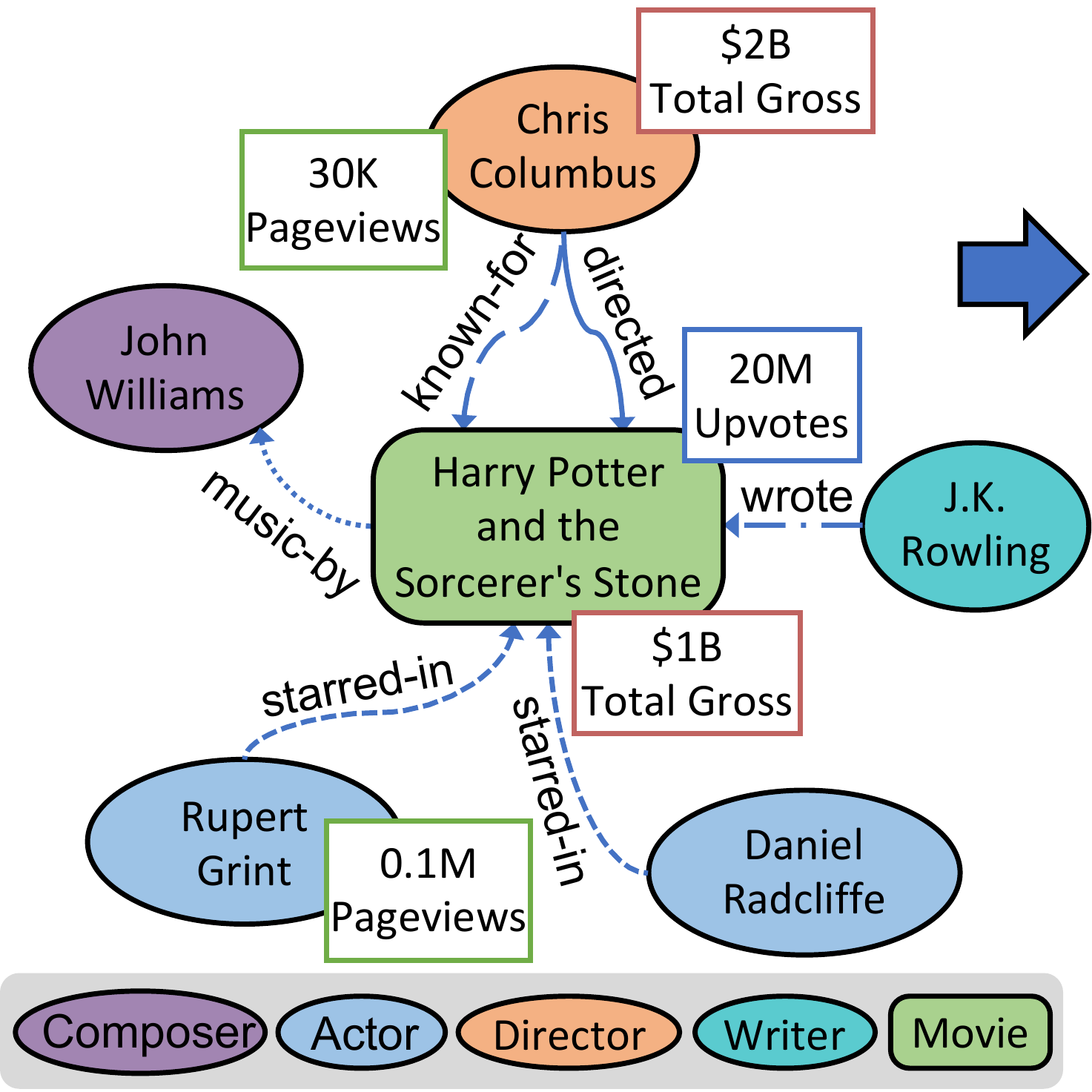}
			\caption{A movie knowledge graph.}
			\label{fig:kg}
			\vspace{0pt}
		\end{subfigure}%
		\begin{subfigure}[b]{.25\textwidth}
			\centering
			\includegraphics[width=\textwidth]{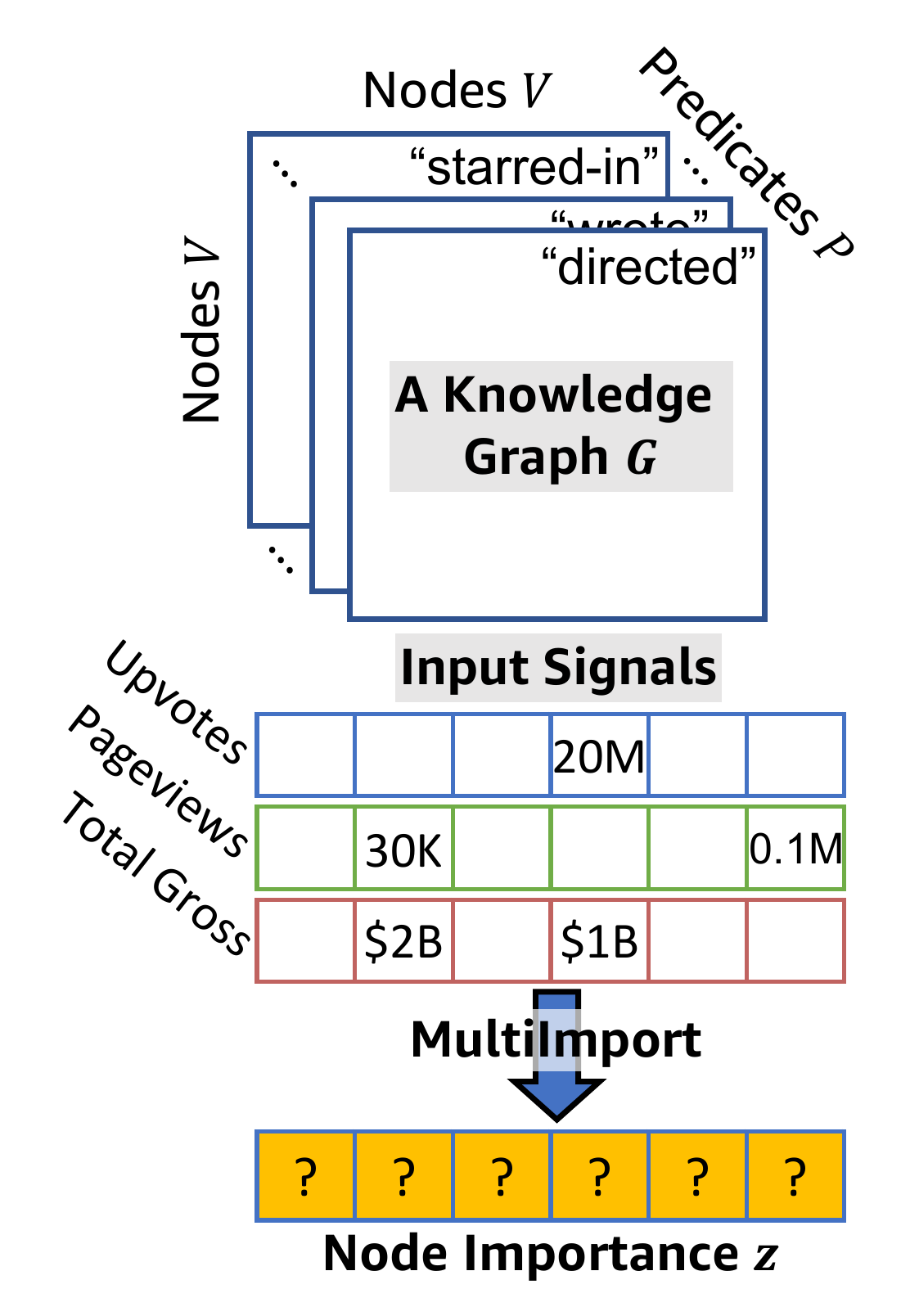}
			\caption{Problem setup.}
			\label{fig:setup}
			\vspace{2pt}
		\end{subfigure}%
	}
	\usebox{\bigimage}\hfill
	\begin{minipage}[b][\ht\bigimage][s]{.40\textwidth}
		\begin{subfigure}{\textwidth}
			\centering
			\includegraphics[width=.99\linewidth]{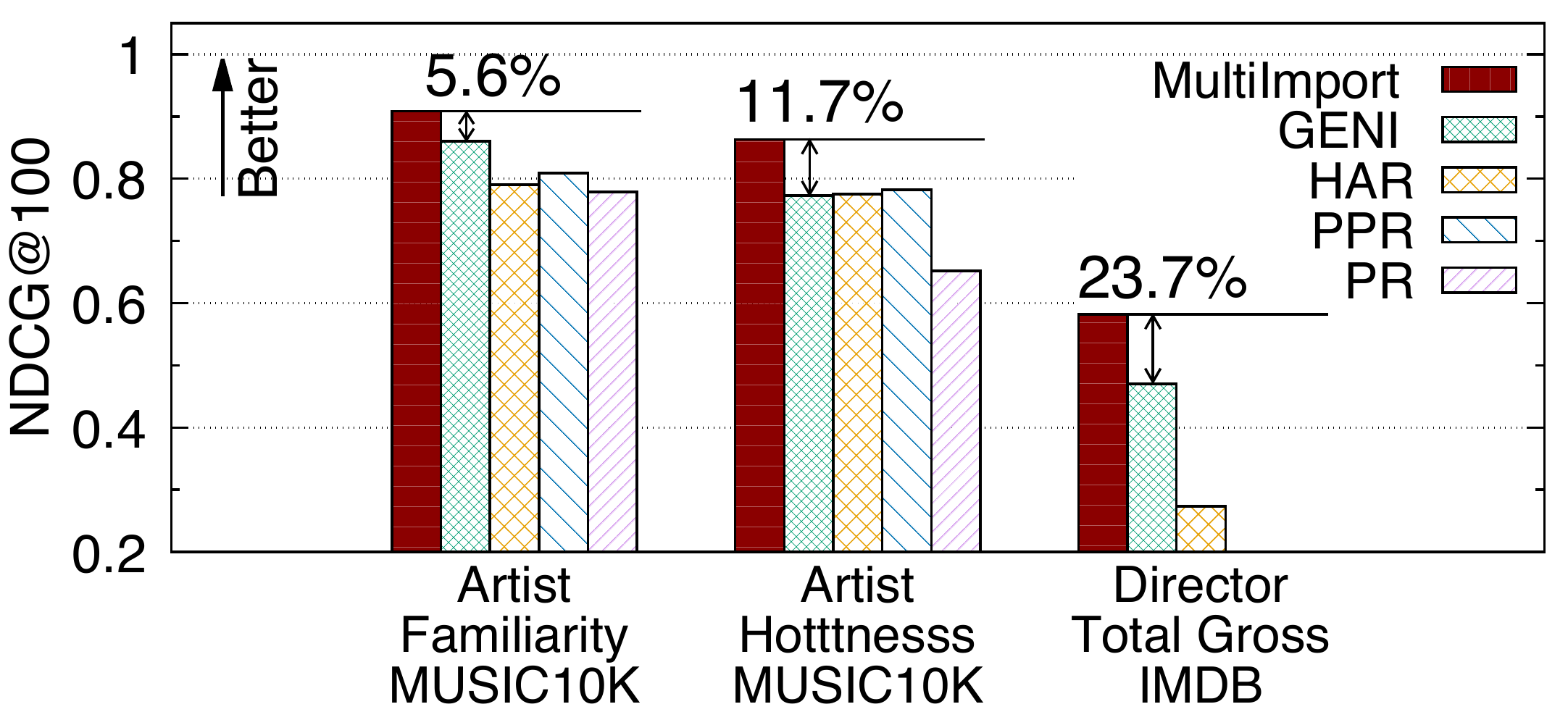}
			\setlength{\abovecaptionskip}{0pt}
			\caption{Accuracy of estimated node importance on two KGs.}
			\label{fig:crownjewel:reconstruction}
		\end{subfigure}%
		\vfill
		\begin{subfigure}{\textwidth}
			\centering
			\includegraphics[width=.99\linewidth]{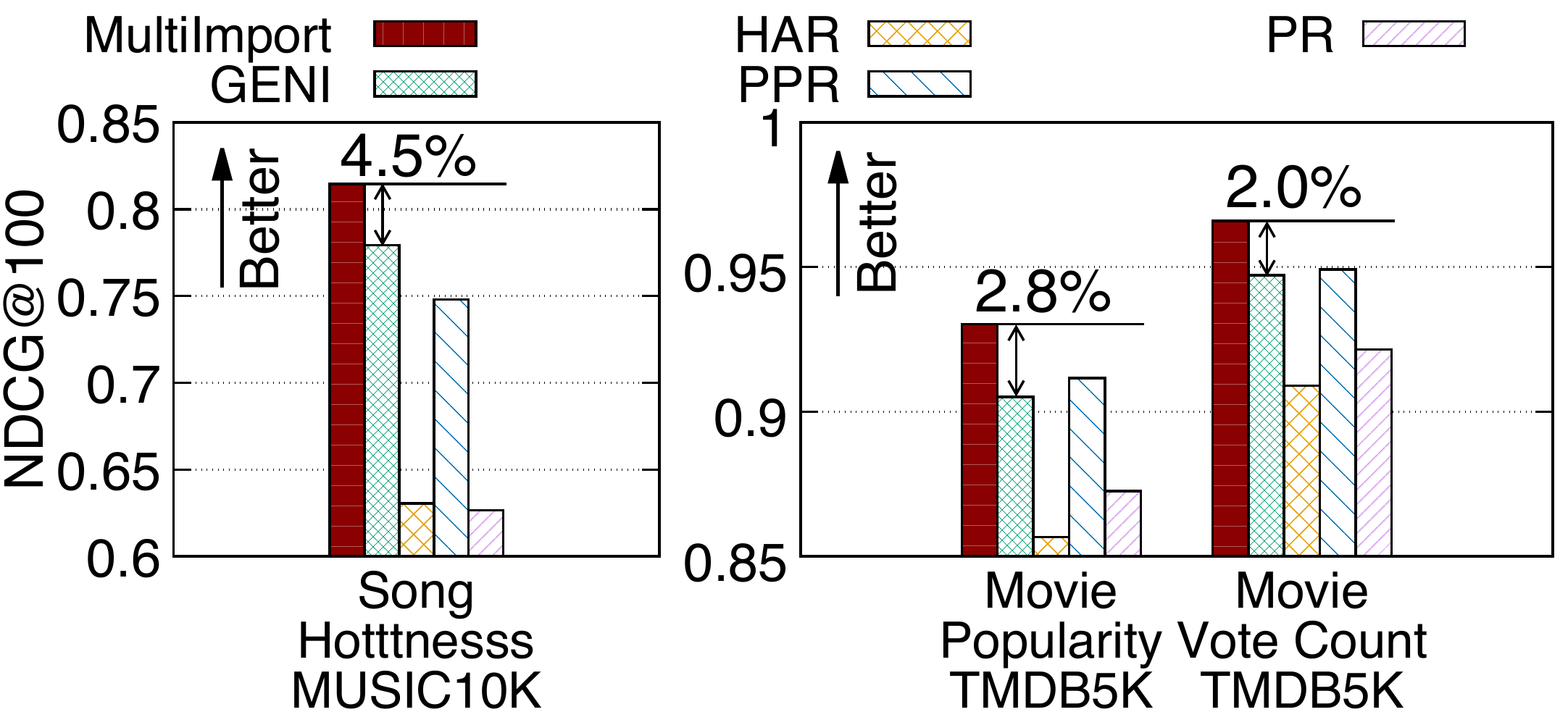}
			\setlength{\abovecaptionskip}{0pt}
			\caption{
				\Signal forecasting performance on two KGs.
			}
			\label{fig:crownjewel:forecasting}
		\end{subfigure}
		\vspace{0pt}
	\end{minipage}
	\setlength{\abovecaptionskip}{1pt}
	\caption{
		(a) A knowledge graph (KG) on a movie and related entities. 
		Node color denotes an entity type and an edge type denotes the type of relation between entities.
		Rectangles represent \signals (e.g., total gross). Note that a single entity can have a variable number of \signals.
		(b) Given a KG and \signals, \method infers the importance of all nodes.
		(c) \method infers up to 23.7\% more accurate node importance than the state-of-the-art method (GENI).
		(d) \method achieves up to 4.5\% higher forecasting results than baselines, with an \ndcg of 0.98 on \tmdb.
		See \Cref{sec:exp} for details.
	}
	\vspace{-1.0em}
\end{figure*}

Real-world networks consist of several types of entities, interacting with each other via multiple types of relations.
These complex and rich interactions between entities from diverse domains are abstracted by a \textit{knowledge graph} (KG),
which is a multi-relational graph where nodes are real-world entities or concepts, and 
edges denote the corresponding relation (also called \textit{predicate}) between nodes.
Given a KG, estimating node importance is a crucial task that has been studied extensively~\cite{page1999pagerank,DBLP:conf/www/Haveliwala02,DBLP:journals/jacm/Kleinberg99,DBLP:journals/kais/TongFP08,DBLP:conf/sdm/LiNY12,DBLP:conf/sigmod/JungPSK17,DBLP:conf/kdd/ParkKDZF19}, 
as it enables a large number of applications such as recommendation, search, and ranking, to name a few.

A key challenge to achieve this goal lies in effectively using input from different sources.
On the one hand, KGs represent how entities are related to each other.
In particular, compared to the conventional graphs that make no distinction between edges,
KGs provide abundant information as they comprise heterogeneous entities and predicates.
For instance, consider the cross-domain KG on the movie ``Harry Potter and the Sorcerer's Stone'' and related entities (\Cref{fig:kg}),
which consists of entities from various domains (e.g., ``actor'', ``composer'', and ``movie'') and multiple relations (e.g., ``directed'' and ``wrote'').

On the other hand, we can often obtain relevant data on the entities in a KG from external sources such as the World Wide Web.
Among them, some are direct indicators of the importance or popularity of an entity
as they capture how much time, money, or attention people have spent on it.
The number of votes and pageviews are examples of such signals.
We call this data that captures node importance \textit{\signal}.
A number of \signals are usually available for a KG, such as the total gross of movies in a movie KG,
although they are often sparse and some are applicable to only specific types of entities, 
as illustrated in \Cref{fig:kg}. 

Existing approaches for measuring node importance include random walk-based methods, such as 
PageRank (\pr)~\cite{page1999pagerank}, Personalized PageRank (\ppr)~\cite{DBLP:conf/www/Haveliwala02}, and \har~\cite{DBLP:conf/sdm/LiNY12}, and
more recent supervised techniques, such as \geni~\cite{DBLP:conf/kdd/ParkKDZF19}, which learn to estimate node importance.
In the context of measuring entity importance in a multi-relational graph,
these methods can be compared in terms of what type of input they can use, as \Cref{tab:method_comparison} summarizes.
While \pr can consider only the graph structure, 
the development of more advanced random walk-based techniques enabled considering additional input.
State-of-the-art results on this task have been achieved by \geni,
which is built upon a supervised framework optimized to use both the KG and an external \signal.

However, all existing approaches can only consider up to one \signal,
even though several \sgnls are typically available from diverse sources.
Also, while it is left to the users to decide which signal to use, no guideline has been provided.
Importantly, by ignoring all other signals except for one, 
they lose information that can complement each other and provide more reliable and accurate evidence for node importance when used together.

In this paper, we present \method, a supervised approach 
that makes an effective use of multiple \signals towards learning accurate and trustworthy node importance in a KG.
Note that among different types of input listed in \Cref{tab:method_comparison}, 
\signals are the most direct and strongest indicator of node popularity.
However, utilizing multiple \sgnls raises several challenges that require careful design choices. 
First, given sparse, potentially overlapping, multiple \signals, 
it is not clear how unknown node importance can be inferred.
Also, using all available \sgnls may lead to worse results, when there exist conflicts among \sgnls.
Developing an effective graph-based estimator is another challenge to model the relation between node importance, \signals, and the KG.

To address these challenges, we model the task using a latent variable model and derive an effective learning objective.
By adopting an iterative clustering-based training scheme, we handle those \sgnls that may deteriorate the estimation quality.
Also, we use predicate-aware, attentive graph neural networks (GNNs) to model the interactions among \signals and the KG.
Our contributions are summarized as follows:
\vspace{-0.25em}
\begin{itemize}
	\item \textbf{Problem Formulation.} 
	We formulate the problem of inferring node importance in a KG from multiple \signals.
	\vspace{-0.25em}
	\item \textbf{Algorithm.} We present \method, a novel supervised method that effectively 
		learns from multiple \signals by handling the aforementioned challenges.
	\item \textbf{Effectiveness.} We show the superiority of \method using experiments on real-world KGs .
	\Cref{fig:crownjewel:reconstruction,fig:crownjewel:forecasting} show that \method outperforms
	existing methods across multiple \sgnls and KGs, achieving up to 23.7\% higher NDCG@100 than the state of the art.
\end{itemize}

\vspace{-0.75em}
\section{Background}
\label{sec:background}

\textbf{Graph Neural Networks} (GNNs)~\cite{DBLP:conf/icml/GilmerSRVD17,DBLP:conf/kdd/YingHCEHL18,gat2018} 
are deep learning architectures for graph-structured data.
GNNs consist of multiple layers, where each one updates the embeddings of each node 
by aggregating the embeddings from the neighborhood, and combining it with the current embeddings.
How the $ \ell $-th layer in GNNs computes the embeddings $ \vect{h}^{\ell}_{i} $ of node $ i $ can be summarized as follows:
\begin{align*}
& \vect{h}^{\ell}_{i} \leftarrow \textsc{Combine}^{\ell} \left( \vect{h}^{\ell-1}_{i}, ~ \textsc{Aggregate}^{\ell} \left( \left\{ \vect{h}^{\ell-1}_j ~ \big| ~ j \in \N(i) \right\} \right) \right)
\end{align*} 
where $ \N(i) $ denotes the neighbors of node $ i $, 
$ \textsc{Aggregate}^{\ell} $ is an operator that aggregates (e.g., averaging) the embeddings of neighbors,
potentially after applying some form of transformation to them, and
$ \textsc{Combine}^{\ell} $ is an operator that merges the aggregated embeddings with the embeddings $ \vect{h}^{\ell-1}_{i} $ of node $ i $.
Different GNNs may use different definitions of $ \N(i) $, $ \textsc{Aggregate}^{\ell}(\cdot) $, and $ \textsc{Combine}^{\ell}(\cdot) $.

\vspace{-0.75em}
\section{Task Description}
\label{sec:probform}
In this section, we present key concepts and the task description.

\textbf{Knowledge Graph.} A \textit{knowledge graph} (KG) is a heterogeneous network with multiple types of entities and relations. 
As shown in \Cref{fig:setup}, a KG can be represented by a third-order tensor $ \kgtensor \in \mathbb{R}^{|V| \times |P| \times |V|} $,
in which a non-zero at $ (s, \rho, o) $ indicates that a subject $ s \in V $ is related to an object $ o \in V $ via a predicate $ \rho \in P$
where $ V $ and $ P $ are the sets of indices for entities and predicates, respectively.
Real-world KGs, such as Freebase~\cite{DBLP:conf/sigmod/BollackerEPST08} and DBpedia~\cite{DBLP:journals/semweb/LehmannIJJKMHMK15}, 
usually contain a large number of predicates.
Also, two entities can be related via multiple predicates as in~\Cref{fig:kg}.

\textbf{Node Feature.}
Node-specific information is often available, and can be encoded in a vector of fixed length $ F $.
Examples include document embedding for the entity description, 
and more domain-specific features like motif gene sets from the Molecular Signatures Database~\cite{subramanian2005gene}.
We use $ \nodefeat \in \mathbb{R}^{|V| \times F} $ to denote all node features.

\textbf{Node Importance.} A \textit{node importance} $ z \in \mathbb{R}_{\ge 0} $ is 
a non-negative real number that represents the importance of an entity in a KG,
with a higher value denoting a higher node importance.
Node importance is a latent quantity, and thus not directly observable.

\textbf{\SIGNAL.} An \textit{\signal} $ S: V' \rightarrow \mathbb{R}_{\ge 0} ~ (V' \subseteq V) $ is a partial map 
between a node and a non-negative real number that represents the significance or popularity of the node. 
For entities in a KG, there are often several external data that could serve as an \signal.
Examples of those \sgnls include the number of copies sold (e.g., of books), 
the total gross of movies and directors, and the number of votes, reviews, and pageviews given for products.
Note that they may highlight node popularity from different perspectives:
e.g., number of clicks in the last one month (higher for trending movies) vs. total number of clicks so far (higher for classic movies).
Also, some signals may be available only for some type of nodes (e.g., the number of tickets sold for movies).

In this work, we consider a set of $ \numS $ \signals $ \{ S_i: {V_i'} \rightarrow \mathbb{R}_{\ge 0} ~|~ V_i' \subseteq V, i = 1, \ldots, \numS \} $
where the \sgnl domain $ V_i' $ might or might not overlap with each other. We note the following facts.
\begin{itemize}[leftmargin=0.9em]
\item \Signals can be in different scales. For example, while signal A ranges from 1 to 5, signal B could range from 0 to 100.
\item \Signals often have high correlation as important nodes tend to have high values across different \sgnls.
However, \sgnls may have a varying degree of correlation 
when \sgnls capture different aspects of node importance or some involve more noise.
\end{itemize}

\textbf{Task Description.}
Based on these concepts, our task of estimating node importance in a KG is summarized as follows 
(\Cref{fig:setup} presents a pictorial overview):
\vspace{-0.25em}
\begin{definition}[Node Importance Estimation]\label{def:problem}
Given a KG $ \kgtensor \in \mathbb{R}^{|V| \times |P| \times |V|} $, node features $ \nodefeat \in \mathbb{R}^{|V| \times F} $, and 
a set of $ \numS $ \signals $ \{ S_i: {V_i'} \rightarrow \mathbb{R}_{\ge 0} ~|~ V_i' \subseteq V, i = 1, \ldots, \numS \} $,
where $ V $ and $ P $ denote the sets of indices of entities and predicates in $ \kgtensor $, respectively,
estimate the latent importance $ z \in \mathbb{R}_{\ge 0}$ of every node in $ V $.
\end{definition}
\vspace{-0.05em}

\vspace{-1.5em}
\section{Methods}
\label{sec:methods}
Inferring node importance in a KG from multiple \signals requires addressing three major challenges.
\begin{enumerate}
\item \textbf{Formulating learning objective.}
Given a KG and potentially overlapping \sgnls, how can we infer latent node importance?
\item \textbf{Handling \rebel \signals.}
Given \signals that may possess different characteristics or involve more noise,
how can we deal with potential conflicts and infer the node importance?
\item \textbf{Effective graph-based estimation.}
How can we effectively model the relations between node importance, \signals, and the KG?
\end{enumerate}
In this section, we present \method that 
addresses the above challenges with the following ideas.
\begin{enumerate}
\item \textbf{Modeling the task using a latent variable model} enables capturing relations between node importance and \signals, 
and provides an optimization framework (\Cref{sec:methods:objective}).
\item \textbf{Iterative clustering-based training} handles \rebel \signals, effectively inferring the node importance (\Cref{sec:methods:rebelsignals}).
\item \textbf{Predicate-aware, attentive GNNs} provide a powerful node importance estimator 
to infer graph-regularized node importance (\Cref{sec:methods:gnn}).
\end{enumerate}
The definition of symbols used in the paper is provided in~\Cref{tab:symbols}.

\begin{figure}[!t]
	\par\vspace{-1em}\par
	\centering
	\setlength{\abovecaptionskip}{1pt}
	\makebox[\linewidth][c]{\includegraphics[width=1.0\linewidth]{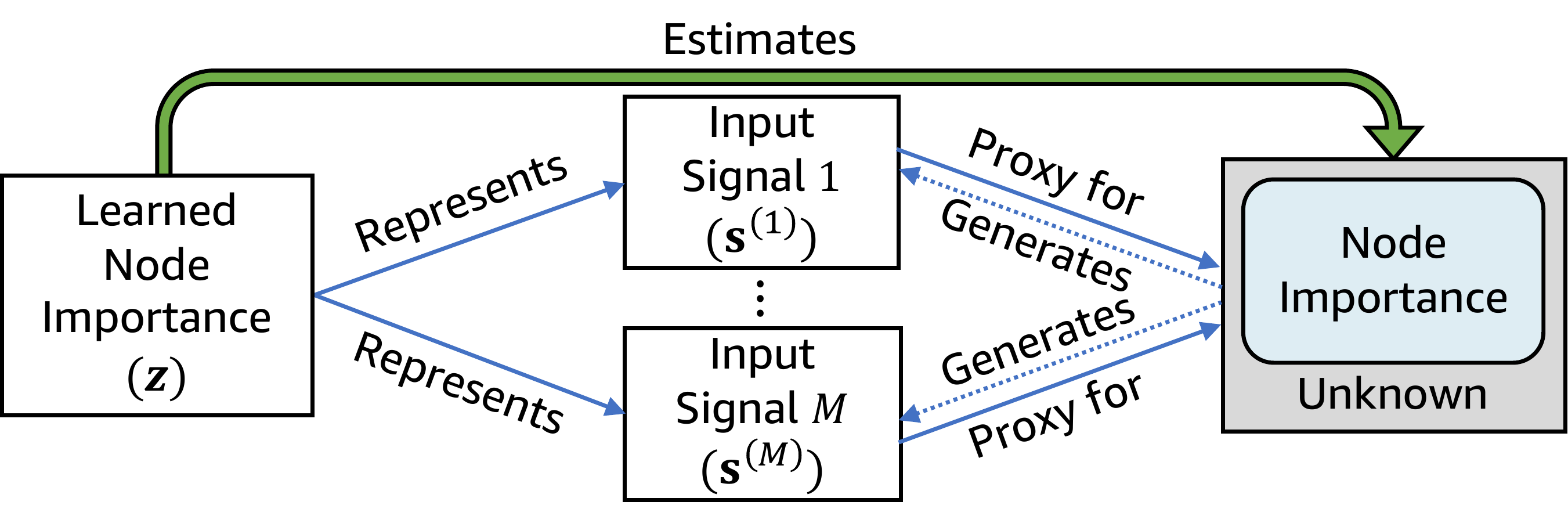}}
	\caption{\method estimates the latent node importance by learning to represent \signals, which are proxies for the unknown node importance.}
	\label{fig:learning_node_importance}
	\vspace{-0.5em}
\end{figure}

\vspace{-1.0em}
\subsection{Learning Objective}
\label{sec:methods:objective}

Given our task to infer node importance, 
one may consider using \signals directly as node features in supervised methods,
as they provide useful cues on the significance of a node.
However, since \sgnls are partially observed,
we will first need to fill in the missing values to use them as a node feature.
Further, even if \signals are available for all nodes,
with all \sgnls treated as node features,
it is not obvious how to infer node importance from them, as node importance is a latent value.

Given that node importance is unknown and cannot be directly observed,
we assume that there is an underlying variable governing node importance, and 
observed \signals are generated by this variable with noise and possibly via non-linear transformations.
Accordingly, we consider an \signal to be a partial indicator of latent node importance, and at the same time,
to be a reasonably good proxy for the unknown node importance.
Based on these assumptions, we approach our goal of estimating unknown node importance by learning to represent \signals.
\Cref{fig:learning_node_importance} illustrates our assumptions on the relationship among the learned node importance, \signals, and node importance.

\textbf{Notations.}
To formally define the learning objective, we introduce a few symbols.
Let $ \kg $ collectively denote the third-order tensor $ \kgtensor $ representing the KG and the node features $ \nodefeat $.
We denote the number of dimensions of vector $ \vect{v} $ by $ \dim(\vect{v}) $.
Given $ \numS $ observed \signals, 
let $ \s^{(i)} $ denote a vector corresponding to the $ i $-th \sgnl, and
$ \setS $ denote a set of $ \numS $ \sgnl vectors, i.e., $ \setS = \{ \s^{(1)},\s^{(2)},\ldots,\s^{(\numS)} \} $.
Let $ \z \in \mathbb{R}^{|V|}$ denote a vector of estimated importance for all nodes.
We use $ \z^{(i)} $ to refer to the vector of estimated importance of those nodes for which \sgnl~$ i $ is available.
Thus, $ \dim(\z^{(i)}) $ equals $ \dim(\s^{(i)}) $.
Since \sgnls are partially observable, we have that $ \dim(\s^{(i)}) = \dim(\z^{(i)}) \le |V| = \dim(\z) $.
To denote the $ j $-th value of a \sgnl or estimated importance,
we use a subscript, e.g., $ s^{(i)}_j $ and $ z^{(i)}_{j} $.

\textbf{Maximum a Posteriori Learning.}
Our goal can be summarized as learning an estimator $ f(\cdot) $
that produces estimated importance $ \z $ for all nodes in the KG.
Specifically, as we consider a graph-based estimator with learnable parameters, 
the estimation by $ f(\cdot) $ is determined by 
the given KG $ \kg $ and its learnable parameters~$ \modelparams $.
In other words, we have:
\begin{align}
\z = f(\kg, \modelparams).
\end{align}

In order to optimize $ f(\cdot) $,
we aim to maximize the posterior probability of model parameters $ \modelparams $ given that we have observed the KG $ \kg $ and \signals $ \setS $:
\begin{align}\label{eq:posterior1}
\max p(\modelparams | \kg, \setS).
\end{align}
By the Bayes' theorem, this is equivalent to maximizing:
\begin{align}\label{eq:posterior2}
p(\modelparams | \kg, \setS) = \frac{p(\kg, \setS|\modelparams) \, p(\modelparams)}{p(\kg, \setS)} 
\propto p(\modelparams) \, p(\kg|\modelparams) \, p(\setS|\kg, \modelparams).
\end{align}

The first term $ p(\modelparams) $ in~\Cref{eq:posterior2} represents the prior probability 
of the model parameters $ \modelparams $,
which we assume to be a Gaussian distribution with zero mean and an isotropic covariance:
\begin{align}\label{eq:prior}
p(\bm{\theta}) = \mathcal{N}(\bm{\theta} | \vect{0}, \lambda^{-1} \mat{I}).
\end{align}

The second term $ p(\kg | \modelparams) $ in~\Cref{eq:posterior2} is the likelihood of 
the observed KG $ \kg $ given $ \modelparams $.
As we later discuss in~\Cref{sec:methods:gnn},
given features $ \vect{x}_i $ of node $ i $,
\method embeds node $ i $ in an intermediate low-dimensional space 
by projecting $ \vect{x}_i $ using a learnable function $ g(\cdot) $.
Let $ (s,\rho,o) $ denote a subject-predicate-object triple in the KG.
Using factorization-based KG embeddings~\cite{DBLP:journals/tkde/WangMWG17}, 
we model the observed triple $ (s,\rho,o) $ as 
a diagonal bilinear interaction between node embeddings $ g(\vect{x}_s) $, $ g(\vect{x}_o) $ 
and the learnable predicate embedding $ \vect{w}_{\rho} $ with normally distributed errors. 
Specifically,
\begin{align}
p(\kg | \modelparams) &= \prod_{(s,\rho,o) \in \kg} p((s,\rho,o)|\modelparams)\\
&= \prod_{(s,\rho,o) \in \kg} \mathcal{N} \left( g(\vect{x}_s)^{\intercal} \diag(\vect{w}_{\rho}) ~ g(\vect{x}_o) \Big| 1, \nu^{-1} \right) \label{eq:kg_likelihood}
\end{align}
where $ \diag(\vect{w}_{\rho}) $ denotes a diagonal matrix where the diagonal entries are given by $ \vect{w}_{\rho} $.
This term can also be seen as an assumption on the homophily between neighboring nodes
in the space represented by node embedding $ g(\cdot) $ and predicate~$ \rho $.

The third term $ p(\setS|\kg, \modelparams) $ of~\Cref{eq:posterior2} is the likelihood of 
observed \signals $ \setS $ given the KG $ \kg $ and model parameters $ \modelparams $.
Given $ \numS $ \sgnls, we assume that they are conditionally independent. Accordingly, we have that:
\begin{align}\label{eq:likelihood1}
p(\setS|\kg, \modelparams) = p(\s^{(1)}|\kg, \modelparams) \cdot \ldots \cdot p(\s^{(\numS)}|\kg, \modelparams).
\end{align}
Recall that $ \z $ is a function of $ \kg $ and $ \modelparams $, or in other words,
$ \kg $ and $ \modelparams $ fully determine $ \z $, and \signals can be partially observed (i.e., $ \dim(\s^{(i)}) $ may not equal $ \dim(\z) $).
\Cref{eq:likelihood1} can be expressed in terms of \signals and the corresponding estimated importance:
\begin{align}\label{eq:likelihood2}
p(\setS|\kg, \modelparams) \propto p(\s^{(1)}|\z^{(1)}) \cdot \ldots \cdot p(\s^{(\numS)}|\z^{(\numS)})
\end{align}
in which the log-likelihood $ \log p(\s^{(i)}|\z^{(i)}) $ of observing \sgnl $ \s^{(i)} $ given $ \z^{(i)} $ is proportional to:
\begin{align}\label{eq:signal_likelihood}
\log p(\s^{(i)}|\z^{(i)}) \propto \log \prod_{j=1}^{\dim(\s^{(i)})} {p (z^{(i)}_j) }^{p(s^{(i)}_j)} = \sum_{j=1}^{\dim(\s^{(i)})} {p(s^{(i)}_j)} \log {p (z^{(i)}_j) }.
\end{align}
Note that taking a negative of \Cref{eq:signal_likelihood} leads to the cross entropy.

\begin{table}[!t]
	\par\vspace{-0.7em}\par
	\small
	\setlength{\abovecaptionskip}{0pt}
	\caption{Table of symbols.}
	\centering
	\makebox[0.4\textwidth][c]{
	\setlength{\tabcolsep}{0.5mm}
	\begin{tabular}{ c | l }
		\toprule
		\textbf{Symbol} & \multicolumn{1}{c}{\textbf{Definition}} \\
		\midrule
		$ \kg $ & knowledge graph with node features \\
		$ V, P $ & set of indices of nodes and predicates in a KG \\
		$ \vect{x}_i, \nodefeat $ & feature vector of node $ i $, and a matrix of all node features \\
		$ \modelparams $ & learnable parameters of the node importance estimator $ f(\cdot) $ \\
		$ \numS $ & number of \signals \\
		$ \s^{(i)} $ & a vector of $ i $-th observed \signal \\
		$ \setS $ & a set of $ \numS $ \signals, i.e., $ \setS = \{ \s^{(1)},\s^{(2)},\ldots,\s^{(\numS)} \} $ \\
		$ \z $ & a vector of estimated importance of all nodes \\
		$ \z^{(i)} $ & a vector of estimated importance of those nodes with \sgnl $ i $  \\
		$ \dim(\vect{z}) $ & number of dimensions of vector $ \vect{z} $ \\
		$ \diag(\vect{w}) $ & a diagonal matrix whose diagonal entries are given by vector $ \vect{w} $ \\
		$ \N(i) $ & neighboring edges of node~$ i $ \\
		$ h^{\ell}_i $ & \makecell[l]{importance of node $ i $ estimated by the $ \ell $-th layer in the GNN} \\
		$ h^{*}_i $ & final estimated importance of node $ i $ by the GNN (i.e., $ z_i = h^{*}_i $) \\
		$ \omega^{\ell}_{ij} $ & \makecell[l]{node $ i $'s attention on the $ m $-th edge from node $ j $ in layer $ \ell $} \\
		$ \pi(\rho_{ij}^{m}) $ & learnable predicate embedding of $ m $-th edge between nodes $ i $ and $ j $ \\
		\bottomrule
	\end{tabular}
	}
	\label{tab:symbols}
\end{table}

With respect to the probability $ p(\s^{(i)}) $ of observing \sgnl $ \s^{(i)} $, we consider two things.
First, \signals can be in different scales. 
Since \sgnls are obtained from diverse sources, 
their values could be in different scales and units, and 
thus may not be directly comparable (e.g., \# clicks vs. dwell time vs. the total revenue in dollars).
Second, for most downstream applications of node importance, the rank of each entity's importance matters much more than the raw value itself.
In light of these observations, we consider the probability of observing a \sgnl vector in terms of ranking.

To do so, once we obtain a list of entity rankings from the \sgnl vector,
we need a probability model that measures the likelihood of the ranked list.
Permutation probability~\cite{DBLP:conf/icml/CaoQLTL07} is one such model,
in which the likelihood of a ranked list is defined with respect to a given permutation of the list,
such that the permutation which corresponds to sorting entities according to their ranking
is most likely to be observed.
In our setting, however, this model is not feasible since there are $ O(|V|!) $ permutations to be considered.
Instead, we use a tractable approximation of it called top one probability.
Given \sgnl $ \s^{(i)} $, the top one probability $ p(s^{(i)}_j) $ of $ j $-th entity in $ \s^{(i)} $ represents 
the probability of that entity to be ranked at the top of the list given the \sgnl values of other entities, and is defined as:
{\setlength{\abovedisplayskip}{1pt}
\setlength{\belowdisplayskip}{2pt}
\begin{align}\label{eq:topone_s}
p(s^{(i)}_j) = \frac{\phi(s^{(i)}_j)}{\sum_{k=1}^{\dim(\s^{(i)})} \phi(s^{(i)}_k)} = 
\frac{\exp(s^{(i)}_j)}{\sum_{k=1}^{\dim(\s^{(i)})} \exp(s^{(i)}_k)}.
\end{align}}Here, $ \phi(\cdot) $ is a strictly increasing positive function, which we define to be an exponential function.
Similarly, given model estimation~$ \z^{(i)} $, the top one probability $ p(z^{(i)}_j) $ of $ j $-th entity in $ \z^{(i)} $ is computed as:
{\setlength{\abovedisplayskip}{1pt}
\setlength{\belowdisplayskip}{1pt}
\begin{align}\label{eq:topone_z}
p(z^{(i)}_j) = \frac{\exp(z^{(i)}_j)}{\sum_{k=1}^{\dim(\z^{(i)})} \exp(z^{(i)}_k)}.
\end{align}}

Taking a negative logarithm of our posterior in~\Cref{eq:posterior2} and plugging in~\Cref{eq:prior,eq:likelihood2,eq:kg_likelihood,eq:signal_likelihood},
we get the following loss:
\begin{tcolorbox}[ams align,boxsep=0pt,boxrule=0pt,left=0pt,right=0pt,top=0pt,bottom=0pt]
\begin{split}\label{eq:loss}
\mathcal{L} =& - \log \left( p(\modelparams) \, p(\kg|\modelparams) \, p(\setS|\kg, \modelparams) \right)\\
=& \left( - \sum_{i=1}^{\numS} \sum_{j=1}^{\dim(\s^{(i)})} p(s^{(i)}_j) \log {p (z^{(i)}_j)} \right)\\
&+ \frac{\nu}{2} \left( \sum_{(s,\rho,o) \in \kg} \left( g(\vect{x}_s)^{\intercal} \diag(\vect{w}_{\rho}) ~ g(\vect{x}_o) - 1 \right)^2 \right) + \frac{\lambda}{2} \lVert \modelparams \rVert^2_2 
\end{split}
\end{tcolorbox}
where $ p(s^{(i)}_j) $ and $ p (z^{(i)}_j) $ are given by~\Cref{eq:topone_s,eq:topone_z}.

\vspace{-1.0em}
\subsection{Handling \Rebel \SIGNALS}
\label{sec:methods:rebelsignals}

In~\Cref{sec:methods:objective}, \method infers node importance from all $ \numS $ \signals.
This is based on the assumption that the given \sgnls have been generated by a common hidden variable as depicted in~\Cref{fig:learning_node_importance},
that is, \signals are homogeneous and a high correlation exists among them.
However, some \sgnls (which we call \textit{\rebel} \sgnls) may exhibit a low correlation with the others
when they possess different characteristics from others, or involve more noise. 
As a result, when given multiple \sgnls where some are weakly correlated with others, 
learning from all of them leads to a worse estimator 
due to the violation of our modeling assumption.

To effectively infer node importance from multiple \sgnls while handling \rebel \sgnls, 
we adopt an iterative cluster\-ing-based training,
where closely related \signals are put into the same cluster and
our estimator is trained using not all the given \sgnls, but only those in the same cluster.
To do so, we need to be able to measure the relatedness of \signals.
Again, considering that \sgnls can be in different scales and ranks are important for downstream applications, 
we compare a pair of \sgnls in terms of the Spearman correlation coefficient, a well-known rank correlation measure.

\begin{figure}[!t]
	\par\vspace{-0.5em}\par
	\centering
	\setlength{\abovecaptionskip}{0pt}
	\makebox[\linewidth][c]{\includegraphics[width=0.99\linewidth]{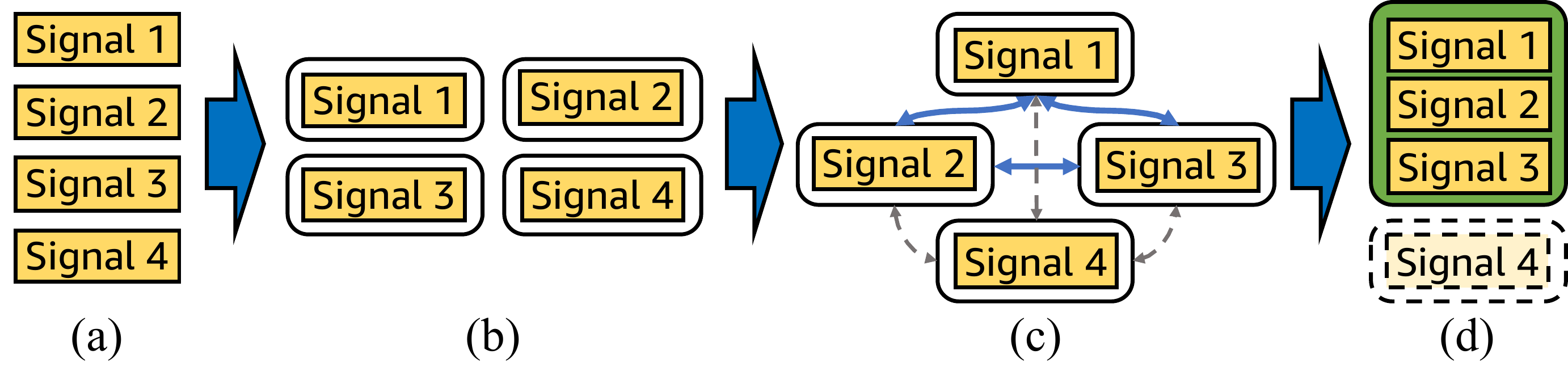}}
	\caption{\method identifies similar \sgnls (those in the green cluster), and infers node importance from them. 
		See text for details of steps (a) to (d).}
	\label{fig:overview}
\end{figure}

However, comparing \signals is not always possible, since they can be disjoint as \sgnls are partially observable.
To handle this, \method (a) initially assigns $ \numS $ \sgnls to their own cluster, 
(b) separately infers node importance from each one,
(c) do a pairwise comparison between observed and inferred values, and
(d) merges clusters by applying existing clustering algorithms, such as DBSCAN~\cite{DBLP:conf/kdd/EsterKSX96}, 
on the pairwise \sgnl similarity.
With the resulting clusters, 
we repeat the same process again until there is no change in the clustering.
This is because learning from an enlarged cluster can lead to a higher modeling accuracy,
as we show in~\Cref{sec:exp:accuracy}.
If there is enough overlap between \sgnl{}s' observed values, we may omit step (b), and
compute their similarity directly from observed values in step (c).
These steps are illustrated in \Cref{fig:overview}, and \Cref{alg:learning} gives our learning algorithm.

Given multiple \sgnls, our focus is to infer a single number for each node, 
which represents the major aspect of node importance, as supported by the \sgnls and relevant to downstream applications.
This requires sorting the final clusters based on their priority.
Examples of such priority policy~$ \pi $ include:
(1) cluster size (prefer a cluster with a larger number of \sgnls);
(2) cluster quality (prefer a cluster with a higher reconstruction accuracy);
(3) signal preference (prefer a cluster with \sgnls that are important for the given application).
In experiments, we use cluster size as our priority.

\textbf{Incremental Learning.} 
Our approach naturally lends itself to incremental learning settings
where new \sgnls are added after the model training.
As in the initial training phase, new \sgnls are first put into their own cluster, and
\method merges them with existing clusters based on the similarity of inferred importance.

\vspace{-1.5em}
\subsection{Graph Neural Networks for\\ Node Importance Estimation}
\label{sec:methods:gnn}
Given this optimization framework,
we now present a supervised estimator $ f(\cdot) $ that can model complex relations among \signals and the information of the KG.
In \method, we utilize GNNs, which have been shown to be a powerful model for learning on a graph.
We adopt and improve upon the recent development of GNNs, such as the dynamic neighborhood aggregation via an attention mechanism,
by making extensions and simplifications.

\begin{algorithm}[!t]
	\small
	\DontPrintSemicolon
	\SetNoFillComment
	\SetKwComment{Comment}{$\triangleright$\ }{}
	\KwIn{knowledge graph $ \kg $, \signals $ \setS $, merge threshold $ t $, priority policy $ \pi $.}
	\KwOut{estimated node importance.}
	
	\Repeat{there is no change in the clustering}{
		Assign \signals without cluster membership (e.g., newly added \sgnls) to their own cluster, if any\\
		
		\ForEach{cluster $c$}{
			Infer node importance by training an estimator $ f_c(\cdot) $ with the loss function in \Cref{eq:loss}
		}
		
		Merge those clusters whose similarity is greater than threshold~$ t $
	}
	\Return node importance inferred from the cluster that has the highest priority according to the policy $ \pi $
	\caption{Learning algorithm}
	\label{alg:learning}
\end{algorithm}

\method first projects features~$ \vect{x}_i $ of node $ i $ to a low-dimen\-sional space using a learnable function $ g(\cdot) $.
Let $ \vect{x}'_i = g(\vect{x}_i) $. 
As discussed in~\Cref{sec:methods:objective}, \method allows assuming homophily among neighboring nodes in this embedding space.
Given these intermediate node embeddings, 
our estimator further transforms them into one dimensional embedding
to directly represent nodes by their importance.
To do so, \method uses another learnable function $ g'(\cdot) $.
In other words, \method represents node $ i $ as $ h^{0}_i \in \mathbb{R} $ in the space of node importance
such that $ {h^{0}_i = g'(\vect{x}'_i)} $.
Note that both $ g(\cdot) $ and $ g'(\cdot) $ are learnable functions, and 
can be a simple linear transformation or multi-layer neural networks.

Then, given $ h^{0}_i $ for all $ i $,
we apply attentive GNNs to it on the given KG to obtain graph-regularized node importance
such that the estimated importance smoothly changes with respect to the KG in a predicate-aware manner.
\method is a multi-layer GNN with $ L $ layers.
The $ \ell $-th layer performs a weighted aggregation of the node importance estimated by the $ (\ell\!-\!1) $-th layer from the neighborhood $ \N(i) $
to produce a new estimation $ h^{\ell}_i $ for node $ i $:
{\setlength{\abovedisplayskip}{1pt}
\setlength{\belowdisplayskip}{1pt}
\begin{align}
h^{\ell}_i &= \sum_{(j, m) \in \N(i)} \omega^{\ell}_{(i,j,m)} ~ h^{\ell-1}_j	
\end{align}}Although attentive GNNs usually compute the attention weight 
for neighboring nodes assuming simple graphs,
KGs are directed graphs with parallel edges.
Therefore, instead of node-level attention, we compute edge-level attention weights, 
which enables making a distinction among edges between two nodes.
We define $ \N(i) $ to be a set of neighboring edges of node~$ i $
such that $ (j, m) \in \N(i) $ if there exists an $ m $-th edge between nodes $ i $ and $ j $ (under some edge ordering).
The weight $ \omega^{\ell}_{(i,j,m)} $ of the edge $ (j, m) \in \N(i) $ is then computed using a predicate-aware attention
parameterized by a weight vector $ \vect{a}_{\ell} $:
{\setlength{\abovedisplayskip}{1pt}
\setlength{\belowdisplayskip}{1pt}
\begin{align}
\omega^{\ell}_{(i,j,m)}\!&=\!\frac{
	\exp \left( \text{LeakyReLU} \left( \vect{a}_{\ell}^\intercal \left[ h^{\ell-1}_i  \big|\big| \pi(\rho_{ij}^{m}) \big|\big| h^{\ell-1}_j \right] \right) \right) 
}{
	\sum_{(k, n) \in \N(i)} \exp \left( \text{LeakyReLU} \left( \vect{a}_{\ell}^\intercal \left[ h^{\ell-1}_i \big|\big| \pi(\rho_{ik}^{n}) \big|\big| h^{\ell-1}_k \right] \right) \right)
}
\end{align}}where $ {\rho}_{ij}^{m} $ denotes the predicate type of $ m $-th edge between nodes~$ i $ and $ j $, 
$ \pi(\cdot) $ is a learnable function that maps a predicate type to its embedding 
(i.e., $\pi(\rho_{ij}^{m})$ is the embedding of the predicate of the $ m $-th edge between nodes $ i $ and $ j $), and
$ || $ is a concatenation operator.
Motivated by \geni, we generate the final estimation $ h^{*}_i $ for node $ i $
by making a centrality-based adjustment to the estimation $ h^{L}_i $ made by the final layer $ L $,
where in-degree $ d_i $ of node $ i $ is used to define its centrality~$ c_i $:
\begin{align}
\begin{split}
c_i &= \alpha \cdot \log(d_i+\epsilon) + \beta\\
h^{*}_i &= \text{ReLU} \left( c_i \cdot h^{L}_i \right)
\end{split}
\end{align}
where $ \alpha $ and $ \beta $ are learnable parameters and $ \epsilon $ is a small positive value.
In summary,  the final estimated node importance $ \vect{z} $ is produced as follows:
\begin{align}
\z = f(\kg, \modelparams) = [ h^{*}_1, \ldots, h^{*}_{|V|} ]^\intercal.
\end{align}

\section{Experiments}
\label{sec:exp}
In this section, we address the following questions.
{\setlist[enumerate]{nosep,leftmargin=1.8em}
\begin{enumerate}[label=\textbf{Q{{\arabic*}}.},ref=Q\arabic*]
	\item \label{sec:exp:q1} \textbf{Accuracy.} 
	How consistent is estimated node importance with \signals?
	In particular, how does the estimation performance change when multiple \signals are considered?
	\item \label{sec:exp:q2} \textbf{Use in downstream tasks.} 
	How useful is the estimated importance for downstream tasks?
	\item \label{sec:exp:q3} \textbf{Handling \rebel \sgnls.} 
	How does a \rebel \sgnl affect the performance, and how well does our method handle it?
\end{enumerate}
After describing datasets, evaluation plans, and baselines in \Cref{sec:exp:dataset,sec:exp:evaluation,sec:exp:baseline},
we address the above questions in \Cref{sec:exp:accuracy,sec:exp:usefulness,sec:exp:rebel_signals}.
Experimental settings are presented in~\Cref{sec:appendix}.
}

\vspace{-1em}
\subsection{Dataset Description}
\label{sec:exp:dataset}

We used four publicly available real-world KGs that have different characteristics, and 
were used in a previous study on node importance estimation~\cite{DBLP:conf/kdd/ParkKDZF19}.
We constructed these datasets following the description in \cite{DBLP:conf/kdd/ParkKDZF19}.
Below we give a brief description of these KGs.
Statistics of these data are given in~\Cref{tab:datasets}, and 
the list of available \signals in each KG is provided in~\Cref{tab:datasets:signals}.

\textbf{\fb}~\cite{DBLP:conf/nips/BordesUGWY13} is a KG sampled from the Freebase knowledge base~\cite{DBLP:conf/sigmod/BollackerEPST08},
which consists of general knowledge harvested from many sources, and compiled by collaborative efforts.
\fb is much denser and contains a much larger number of predicates than other KGs.

\textbf{\music} is a music KG representing the relation between songs, artists, and albums.
\music is constructed from a subset of the Million Song Dataset\footnote{http://millionsongdataset.com/}, and 
it provides three \signals called ``song hotttnesss'', ``artist hotttnesss'', and ``artist familiarity'', 
which are popularity scores computed by considering several relevant data such as playback count.

\textbf{\tmdb} is a movie KG representing relations among movie-related entities such as movies, actors, directors, crews, casts, and companies.
\tmdb is constructed from the TMDb 5000 datasets\footnote{https://www.kaggle.com/tmdb/tmdb-movie-metadata}, and 
contains several \sgnls including the ``popularity'' score computed by considering relevant statistics 
like the number of votes\footnote{https://developers.themoviedb.org/3/getting-started/popularity}.

\textbf{\imdb} is a movie KG constructed from the daily snapshot of the IMDb dataset\footnote{https://www.imdb.com/interfaces/}
on movies and related entities, e.g., genres, directors, casts, and crews.
\imdb is the largest KG among the four KGs.
As IMDb dataset provides only one \signal (\# votes), we collected popularity \sgnl from TMDb for 5\% of the movies in \imdb.

\begin{table}[!t]
\small
\tablefont
\centering
\setlength{\abovecaptionskip}{1pt}
\caption{
Real-world KGs used in our evaluation. 
These KGs vary in different aspects, such as the size and number of predicates.
SCC: Strongly connected component.
}
\centering
\makebox[0.4\textwidth][c]{
\begin{tabular}{ c | r | r | r | r }
	\toprule
	\textbf{Name} & \textbf{\makecell{\# Nodes}} & \textbf{\makecell{\# Edges}} & \textbf{\makecell{\# Predicates}} & \textbf{\makecell{\# SCCs}} \\
	\midrule
	\fb & 14,951 & 592,213 & 1,345 & 9 \\
	\music & 24,830 & 71,846 & 10 & 130 \\
	\tmdb & 123,906 & 532,058 & 22 & 15 \\
	\imdb & 1,567,045 & 14,067,776 & 28 & 1 \\
	\bottomrule
\end{tabular}
}
\label{tab:datasets}
\end{table}

\begin{table}[!t]
\small
\tablefont
\centering
\setlength{\abovecaptionskip}{1pt}
\caption{
	\Signals in real-world KGs. 
	The percentage of nodes covered by each \sgnl is given in the parentheses.
}
\centering
\makebox[0.4\textwidth][c]{
	\setlength{\tabcolsep}{1.0mm}
	\begin{tabular}{ c | c | l }
		\toprule
		\textbf{Name} & \textbf{Type} & \textbf{\makecell{\SIGNALS}} \\
		\midrule
		\fb & Generic & \makecell[l]{\# Pageviews, \# total edits, and \# page watchers\\ on Wikipedia (all 94\%)} \\ \midrule
		\multirow{2}{*}{\music} & Artist & Artist hotttnesss (14\%) and artist familiarity (16\%) \\
		& Song & Song hotttnesss (17\%) \\ \midrule
		\multirow{2}{*}{\tmdb} & Movie & Popularity, revenue, budget, and vote count (all 4\%) \\
		& Director & Box office grosses for top 200 directors \\ \midrule
		\multirow{2}{*}{\imdb} & Movie & \# Votes (14\%) and popularity (from TMDb, 5\%) \\
		& Director & Box office grosses for top 200 directors \\
		\bottomrule
	\end{tabular}
}
\label{tab:datasets:signals}
\end{table}

\begin{table*}[!t]
	\par\vspace{-1.2em}\par
	\small
	\renewcommand{\aboverulesep}{1pt}
	\renewcommand{\belowrulesep}{1pt}
	\setlength{\abovecaptionskip}{1pt}
	\tablefont
	\caption{
		\method estimates node importance more accurately than baselines, and using additional \sgnls improves the accuracy.
		\methodone is the same as \method except that it used only one \sgnl denoted with an asterisk (*).
		Methods that can use only one \signal also used the one marked with an asterisk (*).
		The best result is in bold and in dark gray. The second best result is underlined and in light gray.
		TR: Training. ID: In-Domain. OOD: Out-Of-Domain.
	}
	\centering
	\setlength{\tabcolsep}{1.5mm}
	\begin{subtable}{1.0\textwidth}
		\centering
		\begin{tabular}{ c | c | c | c | c | c | c }
			\toprule
			& \multicolumn{3}{c|}{\textbf{\fb}} & \multicolumn{3}{c}{\textbf{\music}} \\
			\textbf{Method} & \textbf{\makecell{\# Page Watchers*\\(Generic, TR, ID)}} & \textbf{\makecell{\# Total Edits\\(Generic, TR, ID)}} & \textbf{\makecell{\# Pageviews\\(Generic, ID)}} & \textbf{\makecell{Familiarity*\\(Artist, TR, ID)}} & \textbf{\makecell{Hotttnesss\\(Artist, TR, ID)}} & \textbf{\makecell{Hotttnesss\\(Song, OOD)}} \\
			\midrule
			\pr				& 0.7747 $ \pm $ 0.02 & 0.8579 $ \pm $ 0.00 & 0.8441 $ \pm $ 0.00 & 0.7788 $ \pm $ 0.01 & 0.6520 $ \pm $ 0.00 & 0.4846 $ \pm $ 0.00 \\
			\ppr			& 0.7810 $ \pm $ 0.02 & 0.8604 $ \pm $ 0.00 & 0.8450 $ \pm $ 0.00 & 0.8090 $ \pm $ 0.01 & 0.7823 $ \pm $ 0.01 & 0.6422 $ \pm $ 0.02 \\
			\har			& 0.7625 $ \pm $ 0.01 & 0.9080 $ \pm $ 0.00 & 0.8732 $ \pm $ 0.00 & 0.7905 $ \pm $ 0.01 & 0.7751 $ \pm $ 0.01 & 0.6377 $ \pm $ 0.01 \\
			\geni			& 0.8548 $ \pm $ 0.02 & 0.8787 $ \pm $ 0.04 & 0.8464 $ \pm $ 0.03  & 0.8603 $ \pm $ 0.01 & 0.7727 $ \pm $ 0.02 & 0.6804 $ \pm $ 0.01 \\ \midrule
			\methodone		& \secondbest{0.8879 $ \pm $ 0.02} & \secondbest{0.9250 $ \pm $ 0.03} & \secondbest{0.8863 $ \pm $ 0.03} & \secondbest{0.8839 $ \pm $ 0.01} & \secondbest{0.8046 $ \pm $ 0.02} & \secondbest{0.7109 $ \pm $ 0.02} \\
			\textbf{\method} & \best{0.9150 $ \pm $ 0.01} & \best{0.9498 $ \pm $ 0.01} & \best{0.9066 $ \pm $ 0.01} & \best{0.9083 $ \pm $ 0.00} & \best{0.8633 $ \pm $ 0.02} & \best{0.7173 $ \pm $ 0.01} \\
			\bottomrule
		\end{tabular}
	\end{subtable}
	{\tiny \phantom{space}}
	\begin{subtable}{1.0\textwidth}
		\centering
		\begin{tabular}{ c | c | c | c | c | c | c | c }
			\toprule
			& \multicolumn{4}{c|}{\textbf{\tmdb}} & \multicolumn{3}{c}{\textbf{\imdb}} \\
			\textbf{Method} & \textbf{\makecell{Popularity*\\(Movie, TR, ID)}} & \textbf{\makecell{Vote Count\\(Movie, TR, ID)}} & \textbf{\makecell{Revenue\\(Movie, ID)}} & \textbf{\makecell{Total Gross\\(Director, OOD)}} & \textbf{\makecell{\# Votes*\\(Movie, TR, ID)}} & \textbf{\makecell{Popularity\\(Movie, TR, ID)}} & \textbf{\makecell{Total Gross\\(Director, OOD)}} \\
			\midrule
			\pr				& 0.8294 $ \pm $ 0.02 & 0.8482 $ \pm $ 0.00 & 0.9009 $ \pm $ 0.00 & 0.8829 $ \pm $ 0.00 & 0.7927 $ \pm $ 0.02 & 0.7019 $ \pm $ 0.00 & 0.0000 $ \pm $ 0.00 \\
			\ppr			& 0.8585 $ \pm $ 0.01 & 0.9133 $ \pm $ 0.01 & 0.9535 $ \pm $ 0.00 & 0.8691 $ \pm $ 0.02 & 0.7927 $ \pm $ 0.02 & 0.7169 $ \pm $ 0.01 & 0.0000 $ \pm $ 0.00 \\
			\har			& 0.8131 $ \pm $ 0.02 & 0.9350 $ \pm $ 0.00 & 0.9537 $ \pm $ 0.00 & 0.9298 $ \pm $ 0.01 & 0.7976 $ \pm $ 0.02 &  \secondbest{0.7671 $ \pm $ 0.00} & 0.2735 $ \pm $ 0.03 \\
			\geni			& 0.9055 $ \pm $ 0.02 & 0.9367 $ \pm $ 0.00 & 0.9646 $ \pm $ 0.00 & 0.9398 $ \pm $ 0.00 & 0.9367 $ \pm $ 0.00 & 0.7079 $ \pm $ 0.01 & 0.4703 $ \pm $ 0.02 \\ \midrule
			\methodone		& \secondbest{0.9075 $ \pm $ 0.02} & \best{0.9555 $ \pm $ 0.00} & \secondbest{0.9650 $ \pm $ 0.00} & \secondbest{0.9497 $ \pm $ 0.00} & \secondbest{0.9493 $ \pm $ 0.00} & 0.7581 $ \pm $ 0.01 & \secondbest{0.5607 $ \pm $ 0.01} \\
			\textbf{\method}& \best{0.9302 $ \pm $ 0.01} & \secondbest{0.9536 $ \pm $ 0.00} & \best{0.9716 $ \pm $ 0.00} & \best{0.9558	$ \pm $ 0.00} & \best{0.9542 $ \pm $ 0.01} & \best{0.8444 $ \pm $ 0.02} & \best{0.5819 $ \pm $ 0.03} \\
			\bottomrule
		\end{tabular}
	\end{subtable}
	\vspace{-1.4em}
	\label{tab:results:accuracy}
\end{table*}

\vspace{-1em}
\subsection{Performance Evaluation}
\label{sec:exp:evaluation}

For evaluation, we use normalized discounted cumulative gain (NDCG),
which is a widely used metric for relevance ranking problems.
Given a list of nodes for which we have the estimated importance and the ground truth scores,
we sort the list by the estimated importance, and consider the ground truth \sgnl value at position~$ i $ (denoted by $ r_i $) 
to compute the discounted cumulative gain at position $ k $ ($ DCG@k $) as follows:
\begin{align*}
DCG@k = \sum_{i=1}^{k} \frac{r_i}{\log_2(i + 1)}.
\end{align*}
The gain is accumulated from the top to the bottom of the list, and gets reduced at lower positions
due to the logarithmic reduction factor.
An ideal DCG at rank position $ k $ ($ IDCG@k $) can be obtained
by sorting nodes by their ground truth values, and computing $ DCG@k $ for this ordered list.
Normalized DCG at position $ k $ ($ NDCG@k $) ranges from 0 to 1, and is defined as:
\begin{align*}
	NDCG@k = \frac{DCG@k}{IDCG@k}
\end{align*}
with higher values indicating a better ranking quality.

In computing NDCG, we consider all entities if the \signal is generic and applies to all entity types as in \fb.
Otherwise, we consider only those entities of the type for which the ground truth signal applies.
For instance, to compute NDCG with respect to movie popularity, we only consider movies.
In general, we compute NDCG over those entities that have ground truth values.
An exception is the director \sgnl, which is limited to the top 200 directors based on their worldwide box office grosses.
When computing NDCG for the director \sgnl, we consider all director entities, and 
assume the ground truth of the directors to be 0 if they are outside the top 200 list.
In our experiments, we report \ndcg; using different values for the threshold $ k $ yielded similar results.

\begin{figure}[!t]
	\par\vspace{-0.5em}\par
	\centering
	\makebox[\linewidth][c]{\includegraphics[width=0.95\linewidth]{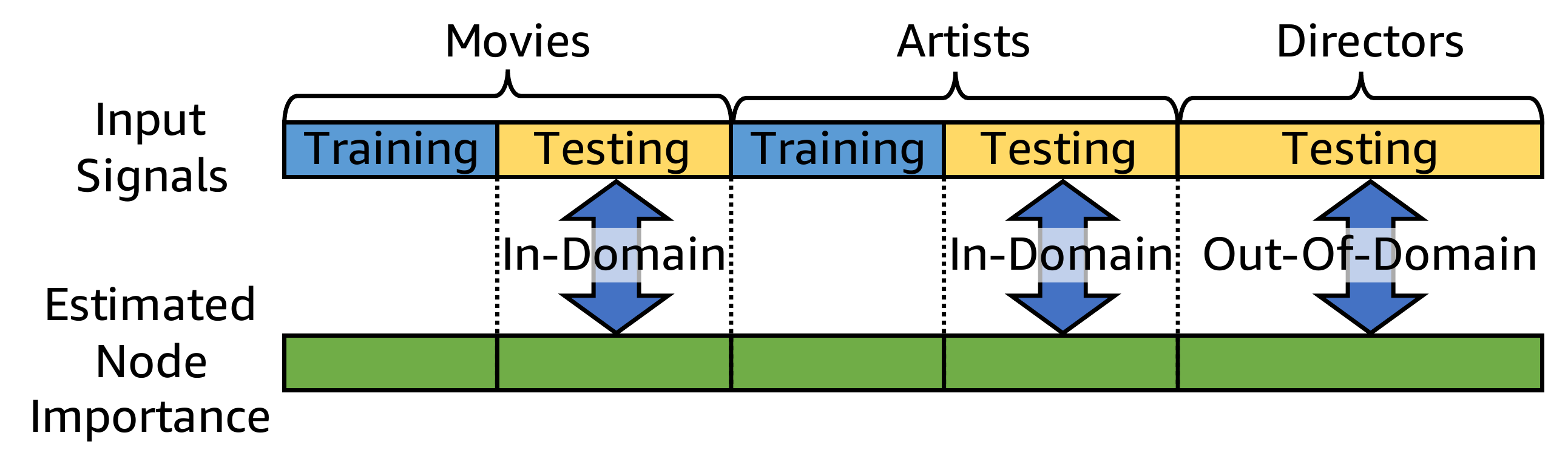}}%
	\setlength{\abovecaptionskip}{-0.2em}
	\caption{In- and out-of-domain evaluation where three \signals are given and two are used for training.}
	\label{fig:evaluation}
	\vspace{-0.5em}
\end{figure}

\textbf{Generalization.}
In order to evaluate the generalization ability of each method,
we perform 5-fold cross validation where 
80\% of the \signals are used for training, 
while the remaining 20\% are used for testing.
Similar results were observed with different number of folds.
Also, since \signals often apply to a specific type of nodes (e.g., directors),
we also consider how well a method generalizes to the nodes of unseen types.
Consider \Cref{fig:evaluation} for an example.
Given \signals on movies, artists, and directors,
movie and artist signals are used for both training and testing,
while the director signal is used for testing alone.
Here we call an evaluation on movies and artists \textit{in-domain}
as training involved \signals on these types of nodes, and 
an evaluation on directors \textit{out-of-domain} since training used no \signal on this type.
It is desirable to achieve a higher accuracy in both criteria.

\vspace{-1.5em}
\subsection{Baselines}
\label{sec:exp:baseline}

We use the following baselines: PageRank (\pr)~\cite{page1999pagerank}, Personalized PageRank (\ppr)~\cite{DBLP:conf/www/Haveliwala02}, \har~\cite{DBLP:conf/sdm/LiNY12}, and \geni~\cite{DBLP:conf/kdd/ParkKDZF19}.
\pr, \ppr, and \har are representative random walk-based algorithms for measuring node importance.
\geni is a supervised method that achieved the state-of-the-art result.
We omitted results from other supervised algorithms, including linear regression, random forests, dense neural networks, 
and other GNNs, such as GAT~\cite{gat2018} and GCN~\cite{DBLP:conf/iclr/KipfW17}, 
as \geni has been shown to outperform them in our experiments.

\vspace{-1.2em}
\subsection{\ref{sec:exp:q1}. Accuracy}
\label{sec:exp:accuracy}

The quality of estimated node importance can be measured by how well it correlates with the observed \sgnls.
That is, accurately estimated node importance $ \z $ should strongly correlate with \signals.
We measure the degree of correlation between the estimation $ \z $ and \signals using NDCG.
In \Cref{tab:results:accuracy}, we report \ndcg of estimated node importance with respect to each \sgnl in our four KGs.

In the table, only those \sgnls marked with TR were used for training.
For baselines that can accept at most one \signal (\ppr, \har, and \geni), 
we used the \sgnl marked with an asterisk (*) as the training \sgnl.
\methodone is identical to \method except that only one signal (marked with *) is used for training.

Overall, \method consistently outperformed baselines across all \sgnls on four datasets,
in terms of both in-domain (ID) and out-of-domain (OOD) evaluation.
Note that \method inferred node importance by learning from the given multiple \sgnls.
A comparison between \method and \methodone shows that 
learning from multiple \signals led to a performance improvement of up to 11\%.
Even if the training \sgnls were given for the same type of entities (e.g., artists or movies), 
considering multiple \sgnls also improved the performance on OOD entities (e.g., songs or directors).
While \geni performed better than random walk-based methods in most cases, 
it was outperformed by \method due to its inability to consider multiple \sgnls.

\begin{table*}[!t]
	\par\vspace{-1.0em}\par
	\small
	\renewcommand{\aboverulesep}{1pt}
	\renewcommand{\belowrulesep}{1pt}
	\tablefont
	\setlength{\abovecaptionskip}{1pt}
	\caption{
		\method achieves the best \sgnl prediction results (marked in bold and in dark gray), where
		we use \signals and estimated node importance as input features to predict another \signal.
		In only one exception, \method achieves the second best result (underlined and in light gray).
		$ \bm{z}_{m} $ denotes node importance estimated by method $ m $.
	}
	\centering
	\setlength{\tabcolsep}{1.00mm}
	\makebox[1.0\textwidth][c]{
		\begin{subtable}{0.33\textwidth}
			\setlength{\abovecaptionskip}{1pt}
			\caption{Predicting ``budget'' on \tmdb by using popularity ($ \bm{p} $), vote count ($ \bm{v} $) and estimated node importance ($ \bm{z} $) as features.}
			\centering
			\begin{tabular}{ c | c | c }
				\toprule
				\textbf{\makecell{Input Features}} & \textbf{\makecell{NDCG@10}} & \textbf{\makecell{NDCG@100}}  \\
				\midrule
				$ \bm{p}, \bm{v} $					& 0.8296 $ \pm $ 0.09 & 0.8176 $ \pm $ 0.00 \\
				$ \bm{p}, \bm{v}, \bm{z}_{\pr} $		& 0.8700 $ \pm $ 0.06 & 0.8260 $ \pm $ 0.00 \\
				$ \bm{p}, \bm{v}, \bm{z}_{\ppr} $		& 0.8602 $ \pm $ 0.07 & 0.8312 $ \pm $ 0.01 \\
				$ \bm{p}, \bm{v}, \bm{z}_{\har} $		& 0.8341 $ \pm $ 0.08 & 0.8062 $ \pm $ 0.02 \\
				$ \bm{p}, \bm{v}, \bm{z}_{\geni} $		& \secondbest{0.8791 $ \pm $ 0.06} & \secondbest{0.8740 $ \pm $ 0.01} \\ \midrule
				$ \bm{p}, \bm{v}, \bm{z}_{\textbf{\method}} $	& \best{0.8930 $ \pm $ 0.07} & \best{0.8757 $ \pm $ 0.00} \\
				\bottomrule
			\end{tabular}
			\label{tab:results:signalprediction:tmdb5k}
		\end{subtable}
		\quad
		\begin{subtable}{0.33\textwidth}
			\setlength{\abovecaptionskip}{1pt}
			\caption{Predicting ``artist hotttnesss'' on \music by using artist familiarity ($ \bm{f} $) and estimated node importance ($ \bm{z} $) as features.}
			\centering
			\begin{tabular}{ c | c | c }
				\toprule
				\textbf{\makecell{Input Features}} & \textbf{\makecell{NDCG@10}} & \textbf{\makecell{NDCG@100}}  \\
				\midrule
				$ \bm{f} $					& 0.3727 $ \pm $ 0.12 & 0.5751 $ \pm $ 0.05 \\
				$ \bm{f}, \bm{z}_{\pr} $		& \secondbest{0.4092 $ \pm $ 0.03} & \secondbest{0.5839 $ \pm $ 0.01} \\
				$ \bm{f}, \bm{z}_{\ppr} $		& 0.3685 $ \pm $ 0.12 & 0.5727 $ \pm $ 0.05 \\
				$ \bm{f}, \bm{z}_{\har} $		& 0.3727 $ \pm $ 0.12 & 0.5751 $ \pm $ 0.05 \\
				$ \bm{f}, \bm{z}_{\geni} $		& 0.3484 $ \pm $ 0.15 & 0.5650 $ \pm $ 0.06 \\ \midrule
				$ \bm{f}, \bm{z}_{\textbf{\method}} $	& \best{0.4186 $ \pm $ 0.06} & \best{0.5886 $ \pm $ 0.00} \\
				\bottomrule
			\end{tabular}
			\label{tab:results:signalprediction:music10k}
		\end{subtable}
		\quad
		\begin{subtable}{0.33\textwidth}
			\setlength{\abovecaptionskip}{1pt}
			\caption{Predicting ``\# page watchers'' on \fb by using \# num pageviews ($ \bm{p} $), \# num total edits ($ \bm{e} $) and estimated node importance ($ \bm{z} $) as features.}
			\centering
			\begin{tabular}{ c | c | c }
				\toprule
				\textbf{\makecell{Input Features}} & \textbf{\makecell{NDCG@10}} & \textbf{\makecell{NDCG@100}}  \\
				\midrule
				$ \bm{p}, \bm{e} $					& 0.8681 $ \pm $ 0.01 & 0.8859 $ \pm $ 0.02 \\
				$ \bm{p}, \bm{e}, \bm{z}_{\pr} $		& 0.9010 $ \pm $ 0.01 & 0.8975 $ \pm $ 0.02 \\
				$ \bm{p}, \bm{e}, \bm{z}_{\ppr} $		& 0.9010 $ \pm $ 0.01 & 0.8975 $ \pm $ 0.02 \\
				$ \bm{p}, \bm{e}, \bm{z}_{\har} $		& \secondbest{0.9052 $ \pm $ 0.00} & 0.8945 $ \pm $ 0.02 \\
				$ \bm{p}, \bm{e}, \bm{z}_{\geni} $		& 0.8884 $ \pm $ 0.00 & \best{0.9076 $ \pm $ 0.02} \\ \midrule
				$ \bm{p}, \bm{e}, \bm{z}_{\textbf{\method}} $	& \best{0.9084 $ \pm $ 0.00} & \secondbest{0.9062 $ \pm $ 0.02} \\
				\bottomrule
			\end{tabular}
			\label{tab:results:signalprediction:fb15k}
		\end{subtable}
	}
	\vspace{-1em}
	\label{tab:results:signalprediction}
\end{table*}

\begin{table}[!t]
	\small
	\renewcommand{\aboverulesep}{1pt}
	\renewcommand{\belowrulesep}{1pt}
	\tablefont
	\setlength{\abovecaptionskip}{1pt}
	\centering
	\caption{
		\method outperforms all baselines in forecasting \sgnls.
		TR denotes the \sgnls used for training \method; baselines were trained with the one marked with an asterisk (*).
		The bottom table shows how data was split.
	}
	\centering
	\setlength{\tabcolsep}{0.75mm}
	
\makebox[0.4\textwidth][c]{
	\begin{subtable}{0.5\textwidth}	
		\centering
		\begin{tabular}{ c | c | c | c | c }
			\toprule
			& \multicolumn{3}{c|}{\textbf{\tmdb}} & \textbf{\music} \\
			\textbf{Method} & \textbf{\makecell{Popularity*\\(Movie, TR)\\\ndcg}} & \textbf{\makecell{Vote Count\\(Movie, TR)\\\ndcg}} & \textbf{\makecell{Budget\\(Movie)\\\ndcg}} & \textbf{\makecell{Hotttnesss*\\(Song, TR)\\\ndcg}} \\
			\midrule
			\pr			& 0.8726 & 0.9215 & 0.9294 & 0.6266 \\
			\ppr		& \secondbest{0.9116} & \secondbest{0.9492} & 0.9640 & 0.7480 \\
			\har		& 0.8567 & 0.9090 & 0.9265 & 0.6306 \\
			\geni		& 0.9051 & 0.9472 & \secondbest{0.9647} & \secondbest{0.7792} \\ \midrule
			\textbf{\method} & \best{0.9303} & \best{0.9660} & \best{0.9829} & \best{0.8145} \\
			\bottomrule
		\end{tabular}
	\end{subtable}
}

	{\tiny \phantom{space}}
\makebox[0.4\textwidth][c]{
	\begin{subtable}{0.5\textwidth}	
		\centering
		\setlength{\tabcolsep}{0.5mm}
		\begin{tabular}{ c | c | c }
			\toprule
			\textbf{Dataset} & \textbf{\makecell{Training (\# Entities)}} & \textbf{\makecell{Testing (\# Entities)}} \\
			\midrule
			\tmdb	& Movies released until 2013 (4243) & Movies released from 2014 (559) \\
			\music	& Songs released until 2005 (1961) & Songs released from 2006 (751) \\
			\bottomrule
		\end{tabular}
	\end{subtable}
}
	\label{tab:results:forecasting}
\end{table}

\vspace{-1.5em}
\subsection{\ref{sec:exp:q2}. Use in Downstream Tasks}
\label{sec:exp:usefulness}

Estimated importance $ \z $ can be viewed as a summary of KG nodes in terms of \signals. 
This summary can be used as a feature in downstream applications. 
In this section, we evaluate how useful \method's estimation $ \z $ is
in downstream signal prediction and forecasting tasks, as opposed to the estimation learned by baselines.

\textbf{Signal Prediction.}
The signal prediction task is to predict some \signal $ \s^{(i)} $ using a machine learning model (M) that uses other \signals $ \s^{(1)},\ldots,\s^{(i-1)} $ and the estimated node importance $ \z $ as input features.
Here $ \z $ can be generated by different methods. Each method uses only $ \s^{(1)},\ldots,\s^{(i-1)} $, and no other \signals (i.e., $ \s^{(i)} $ is not available during generation of $ \z $).
Once $ \z $ is obtained, we compare the performance of M 
when the input features consist of only $ \s^{(1)},\ldots,\s^{(i-1)} $ vs. 
when they are composed of both $ \s^{(1)},\ldots,\s^{(i-1)} $ and $ \z $.
Also, we compare the prediction performance of M as we use $ \z $ obtained with different methods.
The motivation for this signal prediction test is that estimated importance can be considered as de-noising compression of \signals. Thus a high-quality estimate of importance should be a useful input feature for downstream signal prediction model.
We used a linear regression model as our M and optimized it using the loss shown in \Cref{eq:loss} (without the second term),
using \signals from \tmdb, \music, and \fb as input features.
\Cref{tab:results:signalprediction} shows the performance in terms of NDCG.
\method achieved the best results across all datasets, except for one case where it achieved the second best result, 
still obtaining up to 12\% better result compared to when $ \z $ was not used as input features.
This shows that the estimation obtained by \method captures useful information contained in the \signals.

\vspace{-0.1em}
\textbf{Forecasting.} 
The forecasting task is concerned with predicting scores for newly added KG entities.
Consider movie KGs such as \tmdb or \imdb as an example.
Given \signals for movies released until some time point $ t $, 
the task is to accurately estimate the \signals of those movies released after time $ t $.
In this setting, high forecasting accuracy would be an indication of the usefulness of \method.
\Cref{tab:results:forecasting} shows the forecasting performance of \method and baselines on \tmdb and \music, 
and how movies and songs were split into training and testing sets.
We set the split point such that it is close to the end of the range of release dates, while the testing set contains enough number of entities.
For this test, we excluded those movies and songs without the release date. 
On \tmdb, baselines were trained using the \sgnl marked with an asterisk (*), 
while \method used both \sgnls denoted with TR.
Across all \sgnls, \method consistently outperformed baselines,
achieving up to 4.5\% higher \ndcg.

\vspace{-1em}
\subsection{\ref{sec:exp:q3}. Handling \Rebel \Sgnls}
\label{sec:exp:rebel_signals}

\begin{figure}[t!]
	\par\vspace{-1.0em}\par
	\centering
	\setlength{\abovecaptionskip}{5pt}
	\makebox[\linewidth][c]{
	\begin{subfigure}[t]{0.25\textwidth}
		\centering
		\includegraphics[width=\textwidth]{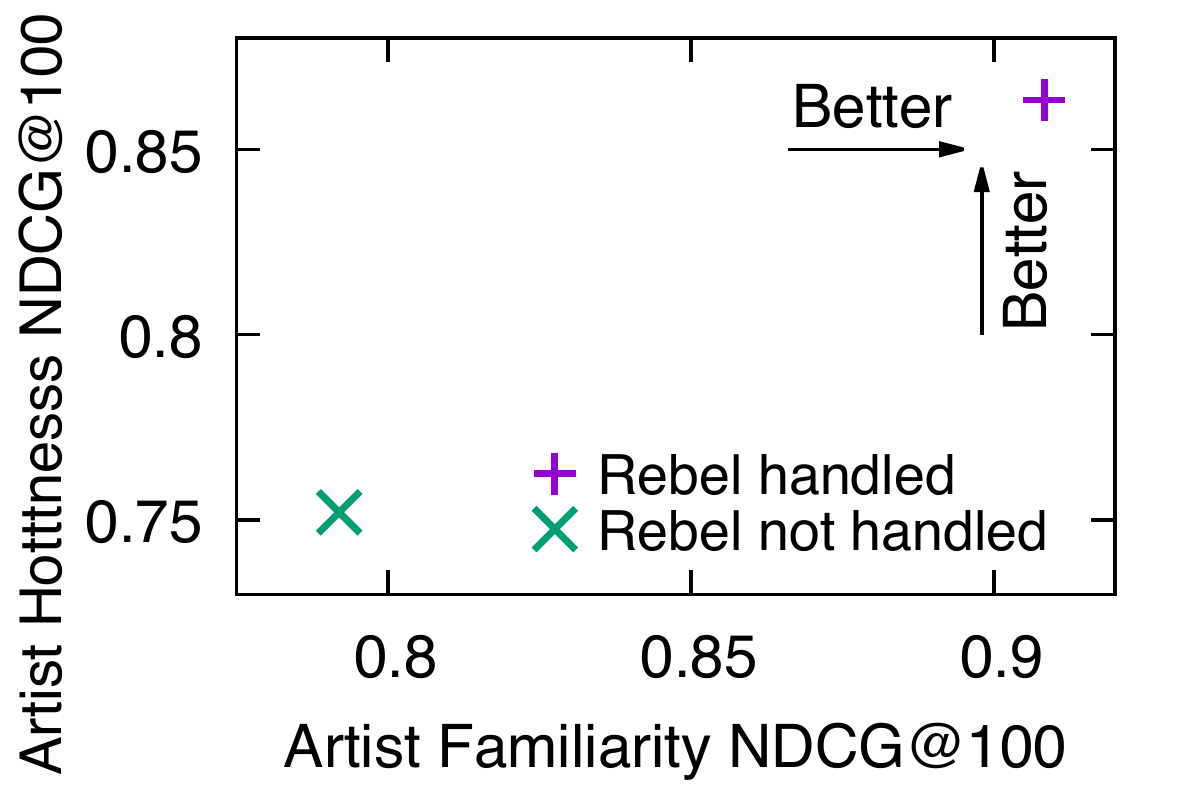}
	\end{subfigure}
	~ 
	\begin{subfigure}[t]{0.25\textwidth}
		\centering
		\includegraphics[width=\textwidth]{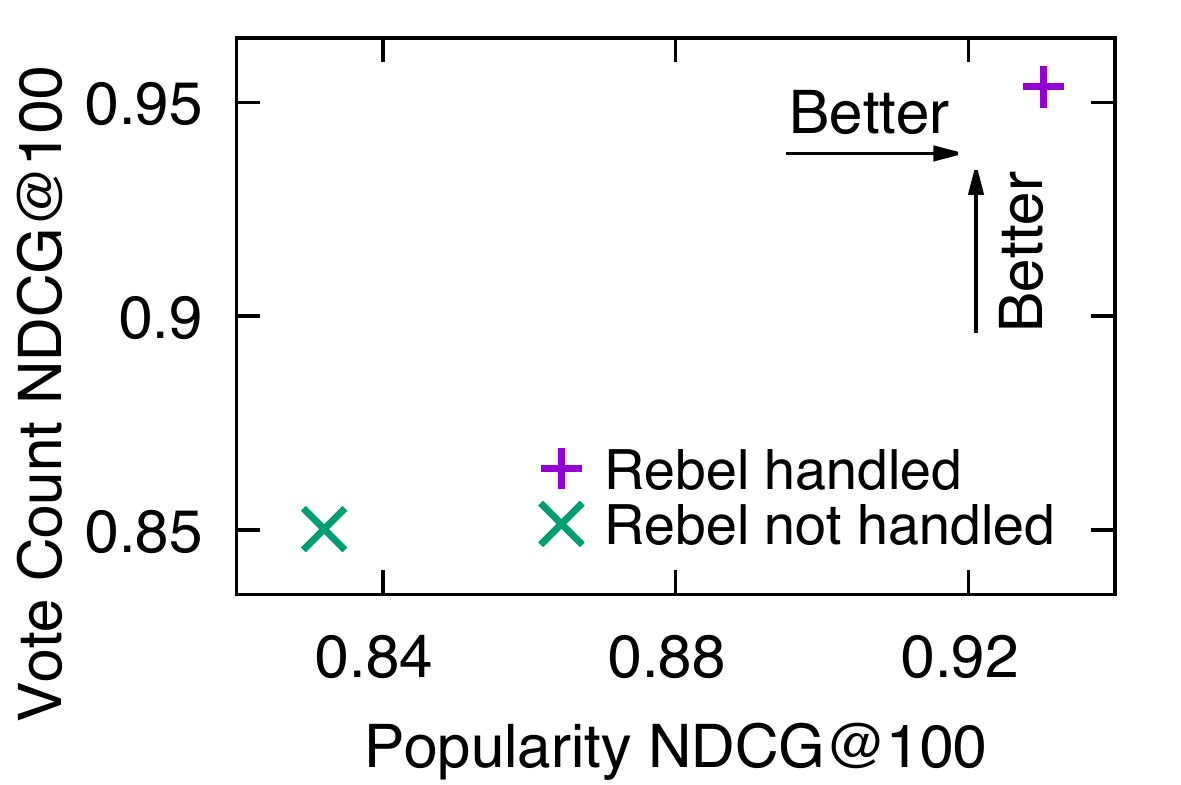}
	\end{subfigure}
	}
	\caption{\Rebel \sgnls can hurt. \method effectively handles them, achieving a higher accuracy. 
	}
	\label{fig:results:rebelsignal}
\end{figure}

To see the effect of handling \rebel \sgnls, we report in~\Cref{fig:results:rebelsignal} 
how the modeling accuracy changes on \music (left) and \tmdb (right) as \rebel \sgnls are handled or not.
For each KG, we trained \method using the two \sgnls shown on the x- and y-axis, 
first time handling \rebel \sgnls and second time ignoring to handle \rebel \sgnls,
and report the \ndcg for both \sgnls.
For \music, we used \pr scores, which turn out to be a weak estimator of node importance in this KG, 
and randomly generated values as \rebel \sgnls;
for \tmdb, we included vote average as a \rebel \sgnl.
In both KGs, by identifying and dropping \rebel \sgnls, \method can achieve up to 14.7\% higher \ndcg in modeling the \signals
in comparison to when failing to handling \rebel \sgnls.

\vspace{-1em}
\section{Related Work}
\label{sec:related}
\textbf{Estimating Node Importance}
is a major graph mining problem, and importance scores have been used in many real world applications, 
including search and recommendation. PageRank (\pr)~\cite{page1999pagerank} is a random walk model 
that propagates the importance of each node by either traversing the graph structure or 
teleporting to a random node with a fixed probability. 
\pr outputs universal node importance for all nodes but these scores do not directly capture proximity between nodes. 
Personalized PageRank (\ppr)~\cite{DBLP:conf/www/Haveliwala02} and 
Random Walk with Restart (RWR)~\cite{DBLP:journals/kais/TongFP08} were then proposed 
to address this limitation by introducing a random jump biased to a particular set of target nodes. 
However, all these methods were designed for homogeneous graphs and thus do not take into account 
different types of edges in a KG. 
To leverage edge type information for node importance estimation, 
\har~\cite{DBLP:conf/sdm/LiNY12} employs a signal propagation schema that is sensitive to edge types.
Recently, with the advent of deep learning on graphs, a graph neural network-based model, 
\geni~\cite{DBLP:conf/kdd/ParkKDZF19}, was developed. 
By using graph attention mechanism, \geni adaptively fuses information from different edge types, and 
achieved the state-of-the-art results for node importance estimation. 
Despite its success, \geni can use only one \sgnl, while multiple \sgnls may come from different sources.
We propose a new method \method that can harmonize \sgnls from multiple sources.

\textbf{Data Fusion and Reconstruction} are related to estimating node importance from \signals.
Data fusion approaches~\cite{DBLP:journals/tkde/YinHY08,DBLP:journals/pvldb/DongBS09,DBLP:journals/pvldb/DongBHS10}
integrate potentially conflicting information about entities from multiple data sources (e.g., websites)
by considering the trustworthiness of data sources and the dependence between them.
Among several values of an object, these methods aim to identify the true value of the object.
Our problem setup is related to, but different from data fusion in that 
our goal is to learn latent node importance which is broadly consistent with \signals,
while focusing on a subset of \sgnls which correlate well with each other,
instead of identifying which \sgnl value is more accurate than others.

Data reconstruction methods aim to complete missing values in the partially observed data.
Matrix and tensor decomposition~\cite{DBLP:journals/siamrev/KoldaB09,DBLP:conf/cikm/ParkJLK16,DBLP:conf/icde/ParkOK17,DBLP:journals/vldb/ParkOK19,DBLP:journals/tpds/OhPJSK19}
are representative approaches to this task, which reconstruct observed data using a relatively small number of latent factors.
While a rank-1 decomposition of \signals is analogous to our problem,
\method simultaneously considers various data sources, such as the KG and \signals, 
while handling rebel signals and employing GNNs for effective inference over a KG.

\textbf{Graph Neural Networks} apply deep learning ideas to arbitrary graph structured data. 
These methods have attracted extensive research interest in recent years~\cite{DBLP:conf/iclr/KipfW17,DBLP:conf/icml/GilmerSRVD17,DBLP:conf/kdd/YingHCEHL18,DBLP:conf/icml/XuLTSKJ18,gat2018}.
Kipf and Welling~\cite{DBLP:conf/iclr/KipfW17} proposed a spectral approach, called GCN, 
which employs a localized first-order approximation of graph convolutions. 
More recently, Veli{\v{c}}kovi{\v{c}} et~al.~\cite{gat2018} proposed graph attention networks (GATs) 
to aggregate localized neighbor information based on attention mechanism. 
GAT provides an efficient framework to integrate deep learning into graph mining, and
has been adopted to recommender systems~\cite{DBLP:conf/www/WuZGHWGC19}, 
knowledge graph reasoning~\cite{DBLP:conf/www/ZhangPWCZZBC19}, and graph classification~\cite{DBLP:conf/kdd/LeeRK18}.
While our work is also based on a graph attention architecture, we additionally consider predicates in attention computation, and 
extend this architecture in a multi-task setting for learning node importance in a KG using multiple \sgnls.

\vspace{-0.5em}

\vspace{-1.5em}
\section{Conclusion}
\label{sec:concl}
Estimating node importance in a KG is a crucial task that has received a lot of interest.
A major challenge in successfully achieving this goal is in utilizing multiple types of input effectively.
In particular, \signals provide strong evidence for the popularity of entities in a KG.
In this paper, we develop an end-to-end framework \method that draws on information from both the KG and external \sgnls,
while dealing with challenges arising from the simultaneous use of multiple \signals, 
such as inferring node importance from sparse \sgnls, and potential conflicts among them.
We ran experiments on real-world KGs to show that \method successfully handles these challenges, and consistently outperforms existing approaches.
For future work, we plan to develop a method for modeling the temporal evolution of node importance in a KG.

\vspace{-1.0em}
\bibliographystyle{ACM-Reference-Format}

\clearpage
\appendix
\section{Appendix}
\label{sec:appendix}

In the appendix, we present experimental settings.

\subsection{Experimental Settings}
\label{sec:exp:settings}

\textbf{PageRank (\pr) and Personalized PageRank (\ppr).}
We used NetworkX 2.3's \textit{pagerank\_scipy} function to run \pr and \ppr.
We used the default parameter values set by NetworkX,
including the damping factor of 0.85 for both algorithms.

\textbf{\har.} 
We implemented \har in Python 3.7.
In experiments, we set $ \alpha $ and $ \beta $ to 0.15 and $ \gamma $ to 0.
We ran the algorithm for 30 iterations.
Normalized \signal values were used 
as the probability of entities (as in \ppr).
All relations were assigned an equal probability.
Among hub and authority scores \har computes for each entity, 
we used the maximum of the two values as \har's estimation, and 
we observed similar results when reporting only one type of scores.

\textbf{node2vec.}
We used the reference node2vec implementation\footnote{https://snap.stanford.edu/node2vec/}
to generate node features.
For \music, \fb, and \tmdb, we set the number of dimensions to 64, and
for \imdb, we set it to 128.
For other parameters, we used the default values used by the reference implementation.

\textbf{\geni.}
We implemented \geni using the Deep Graph Library 0.3.1.
We used \geni with two layers, each consisting of four attention heads.
For the scoring network, we used a two-layer fully-connected neural network,
where the number of hidden neurons in the first hidden layer was 75\% of the input feature dimension.
We set the dimensions of predicate embedding to 10.
\geni was trained using Adam optimizer with $ \beta_1=0.9 $, $ \beta_2=0.999 $, 
a learning rate of 0.005, and a weight decay of 0.0005.
We applied ReLU to estimated node importance, 
ELU to node centrality, and Leaky ReLU to unnormalized attention coefficients.

\textbf{\method.}
We implemented \method using the Deep Graph Library 0.3.1.
For a fair comparison with \geni, we used two layers for \method, and each layer contained four attention heads.
The output from each attention head was averaged, and fed into the next layer.
For $ g(\cdot) $, we used a linear transformation with bias, which projects input features to 75\% of the input feature dimension.
For $ g'(\cdot) $, we used another linear transformation with bias.
We set the dimension of predicate embedding to 10.
For training, we used the Adam optimizer with $ \beta_1=0.9, \beta_2=0.999 $, and
a learning rate of 0.005.
We set $ \lambda $ to 0.001.
We set $ \nu $ to 0.0002 on \fb and \imdb,
applying the corresponding loss term in \Cref{eq:loss} to 20\% and 5\% of randomly sampled edges, respectively;
$ \nu $ was set 0.0 on \music and \tmdb.
ELU was applied for node centrality.

\textbf{Early Stopping.}
For \method and \geni, we applied early stopping based 
on the performance on the validation data set (15\% of training data) with a patience of 30, 
and set the maximum number of iterations to 3000.
For testing, we used the model that achieved the best validation performance.

\textbf{\SIGNAL Preprocessing.}
We applied log transformation for all \sgnls as their distribution is highly skewed,
except for those \sgnls available on \music as they are normalized scores.

\end{document}